\documentclass[lettersize,journal]{IEEEtran}
\usepackage{amsmath,amsfonts}
\usepackage{algorithmic}
\usepackage{algorithm}
\usepackage{array}
\usepackage[caption=false,font=normalsize,labelfont=sf,textfont=sf]{subfig}
\usepackage{textcomp}
\usepackage{stfloats}
\usepackage{url}
\usepackage{verbatim}
\usepackage{graphicx}
\usepackage{cite}
\usepackage{soul} 
\usepackage{color, xcolor} 
\usepackage{amssymb}
\usepackage{multirow}
\usepackage{colortbl}
\usepackage{color}
\usepackage{graphicx}
\usepackage{subcaption}
\usepackage{hyperref} 

\definecolor{RoyalBlue}{rgb}{0.0, 0.1, 0.66}
\hypersetup{
colorlinks=true,
linkcolor=blue,
citecolor=RoyalBlue
}

\usepackage{orcidlink}

\usepackage{booktabs} 

\begin{document}
\newsavebox\CBox
\def\textBF#1{\sbox\CBox{#1}\resizebox{\wd\CBox}{\ht\CBox}{\textbf{#1}}}

\title{Learning an Adaptive and View-Invariant Vision Transformer for Real-Time UAV Tracking}

\author{You Wu, Yongxin Li, Mengyuan Liu, Xucheng Wang, Xiangyang Yang, Hengzhou Ye, Dan Zeng, Qijun Zhao, and Shuiwang Li$^*$
\thanks{
*Corresponding author.
}
\thanks{You Wu, Yongxin Li, Mengyuan Liu, Xiangyang Yang, Hengzhou Ye, and Shuiwang Li are with the College of Computer Science and Engineering, Guilin University of Technology, Guilin 541006, China.
Hengzhou Ye, and Shuiwang Li are also with the Guangxi Key Laboratory of Embedded Technology and Intelligent Information Processing, Guilin 541006, China  (e-mail: wuyou@glut.edu.cn; 2120221108@glut.edu.cn; mengyuaner1122@qq.com; xyyang317@163.com;
yehengzhou@glut.edu.cn; lishuiwang0721@163.com).}
\thanks{Xucheng Wang is with the School of Computer Science, Fudan University, Shanghai 200082, China(e-mail: xcwang317@glut.edu.cn).}
\thanks{Dan Zeng is with the School of Artificial Intelligence, Sun Yat-sen University, Zhuhai 510275, China(e-mail: danzeng1990@gmail.com).}
\thanks{Qijun Zhao is with College of Computer Science, Sichuan University, Sichuan 610065, China (e-mail: qjzhao@scu.edu.cn).}
}



\maketitle

\begin{abstract}

Transformer-based models have improved visual tracking, but most still cannot run in real time on resource-limited devices, especially for unmanned aerial vehicle (UAV) tracking. To achieve a better balance between performance and efficiency, we propose AVTrack, an adaptive computation tracking framework that adaptively activates transformer blocks through an Activation Module (AM), which dynamically optimizes the ViT architecture by selectively engaging relevant components. To address extreme viewpoint variations, we propose to learn view-invariant representations via mutual information (MI) maximization. In addition, we propose AVTrack-MD, an enhanced tracker incorporating a novel MI maximization-based multi-teacher knowledge distillation framework. Leveraging multiple off-the-shelf AVTrack models as teachers, we maximize the MI between their aggregated softened features and the corresponding softened feature of the student model, improving the generalization and performance of the student, especially under noisy conditions.  Extensive experiments show that AVTrack-MD achieves performance comparable to AVTrack's performance while reducing model complexity and boosting average tracking speed by over 17\%.
Codes is available at \url{https://github.com/wuyou3474/AVTrack}.
\end{abstract}


\begin{figure*}[t]
\centering
\includegraphics[width=0.75\linewidth]{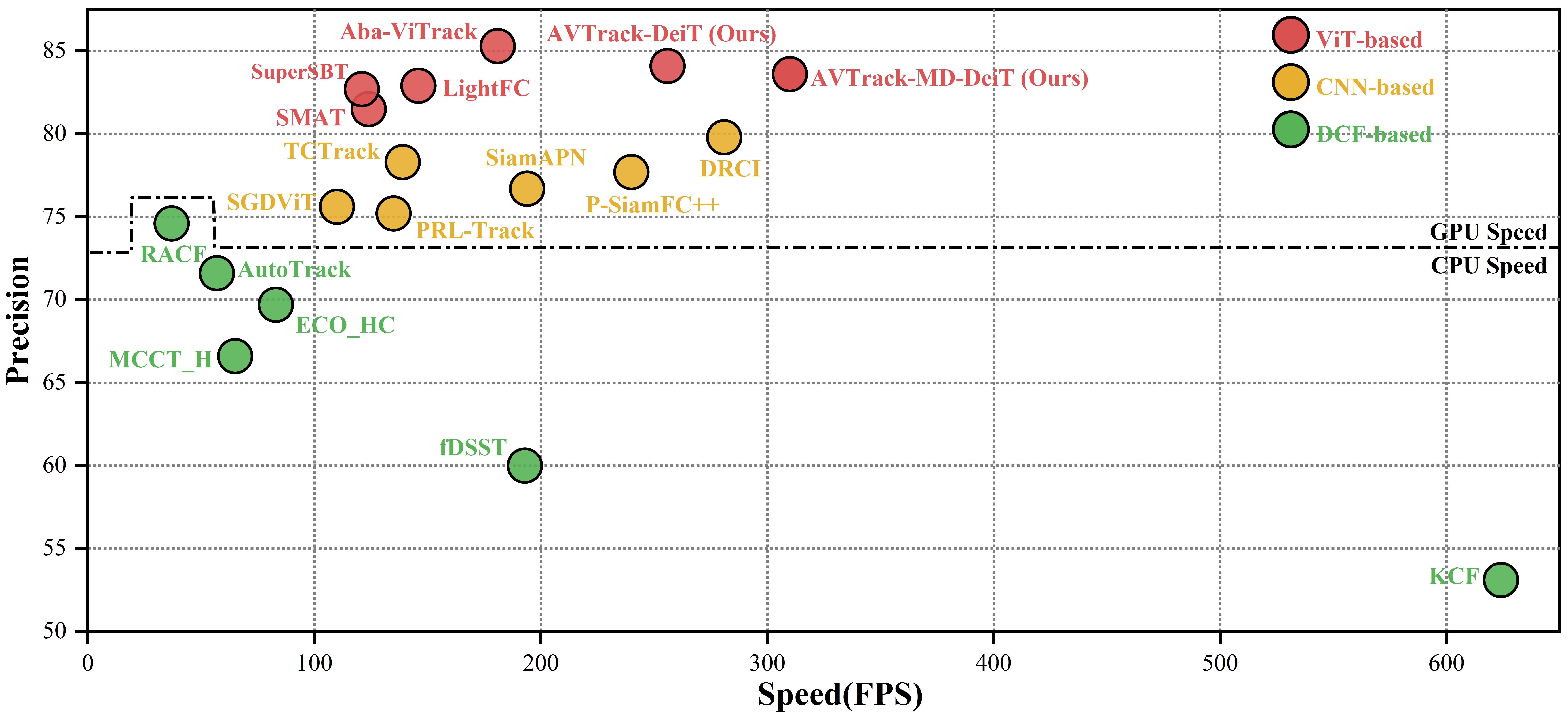}
\caption{Compared to SOTA lightweight trackers, our AVTrack-DeiT sets a new record with 84.1\% precision and a speed of 256.8 FPS. Our AVTrack-MD-DeiT strikes a better trade-off with 83.6\% precision and a speed of about 310.6 FPS. Note that the evaluation was conducted on a PC equipped with an i9-10850K processor (3.6GHz), 16GB of RAM, and an NVIDIA TitanX GPU.}
\label{fig_Prec_Speed}
\end{figure*}

\section{Introduction}

\IEEEPARstart{U}{nmanned} aerial vehicle (UAV) tracking has become increasingly critical due to its diverse applications, including path planning, public safety, visual surveillance, environmental monitoring,  and agriculture, etc.
Specifically, UAV tracking involves determining and predicting the location of a specific object in successive aerial images, which are captured by mobile cameras at relatively high altitudes.
As a result, common challenging scenarios have emerged in the UAV tracking community, including fast object or UAV motion, extreme viewing angles, motion blur, low resolution, and significant occlusions where objects are obscured by other elements.
Moreover, the efficiency of tracking algorithms is crucial, as mobile platforms like UAVs have limited computational resources \cite{li2023adaptive,li2020autotrack,cao2021hift,wu2024learning}.
In order to meet the unique demands of UAV applications, a practical tracker should ensure accurate tracking while running with relatively low computational and power consumption.

Neither discriminative correlation filter (DCF)-based trackers~\cite{henriques2015high,danelljan2015learning,Huang2019LearningAR,li2020autotrack,lin2020learning,kiani2017learning,dai2019visual} nor deep learning (DL)-based trackers~\cite{bertinetto2016fully,li2018high,chen2020siamese,cao2021hift,cao2022tctrack,liu2023bactrack,yuan2024multi,guo2023vit,tang2023learning,zhou2021object,he2024target} can achieve a satisfactory balance between performance and efficiency.  
DCF-based trackers offer higher efficiency but struggle with tracking accuracy and robustness, making them unsuitable for complex UAV tracking scenarios, while DL-based trackers are hindered by high computational costs and sluggish performance, limiting their practical applicability.
Single-stream architecture has recently emerged as a popular strategy in DL-based trackers, seamlessly integrating feature extraction and fusion with pre-trained Vision Transformer (ViT) backbone networks.
A number of representative methods \cite{ye2022joint,chen2022backbone,cui2022mixformer,wu2023dropmae,kou2024zoomtrack,xie2024autoregressive,shi2024evptrack,cai2024hiptrack,wei2023autoregressive,kang2023exploring}, including OSTrack \cite{ye2022joint}, SimTrack \cite{chen2022backbone}, Mixformer \cite{cui2022mixformer}, DropMAE \cite{wu2023dropmae}, ZoomTrack \cite{kou2024zoomtrack}, AQATrack \cite{xie2024autoregressive}, EVPTrack \cite{shi2024evptrack}, and HIPTrack \cite{cai2024hiptrack}, demonstrate the immense potential of this paradigm shift in tracking task.
Motivated by this, Aba-VTrack \cite{li2023adaptive} proposes an effective and efficient DL-based tracker for real-time UAV tracking based on this framework, utilizing an adaptive and background-aware token calculation technique to minimize inference time.
However, the presence of unstructured access operations with a variable number of tokens results in significant time costs.

In this work, we alleviate the above problems by introducing AVTrack~\cite{lilearning2024}, an adaptive computation framework for real-time UAV tracking that enhances the efficiency of Transformer-based models by integrating an Activation Module (AM) into each block. This module dynamically selects relevant components during the tracking, significantly improving speed while maintaining competitive tracking performance.
As shown in Fig. \ref{fig_AVTrack_overview}  (right), to improve efficiency by reducing computation,  AM utilizes only a slice of tokens from both the target template and the search image as input,
it produces an activation probability that decides whether a transformer block should be activated.
By adaptively trimming the ViT at the block level, the proposed method avoids unnecessary unstructured access operations, thereby reducing time consumption.
The rationale behind this is based on the understanding that semantic features or relationships do not uniformly affect the tracking task across different levels of abstraction. Instead, in practice, their impact is closely tied to the characteristics of the target and the scene in which it appears.
In simple scenes, such as when a target moves against a monochromatic background, efficient tracking can be achieved by utilizing the color contrast between the object and the background.
In such cases, this straightforward feature is sufficient. 
However, in real-world scenarios, there are often numerous distractors, such as background clutter, occlusion, similar objects, and changes in viewpoint.
In these challenging scenarios, trackers need to capture and analyze sufficient semantic features and relationships to successfully track the specific object.
These facts demonstrate the dynamic nature of tracking needs, which are inextricably linked to the individual features of the scene and the target being tracked.
The proposed AM is implemented using just a linear layer followed by a nonlinear activation function, making it a straightforward and effective module.
By customizing the architecture of ViTs to suit the specific requirements of tracking tasks, our approach has the potential to achieve a satisfactory balance between performance and efficiency for UAV tracking.

Additionally, AVTrack~\cite{lilearning2024} introduces a novel approach for learning view-invariant feature representations by maximizing mutual information between backbone features from two different views of the target, thereby enhancing the tracker’s robustness against viewpoint variations.
Specifically, mutual information is a measure that quantifies the dependence or relationship between two variables \cite{steuer2002mutual}.
Mutual information maximization involves enhancing the mutual information between different components or variables within a system and is widely used in various computer vision tasks \cite{liu2022temporal,yang2022learning,Hjelm2019LearningDR}.
However, to the best of our knowledge, the effectiveness of this strategy in UAV tracking has not been extensively explored.
In our work, to obtain view-invariant representations, the tracker learns to preserve essential information about the target regardless of viewpoint changes by maximizing the mutual information between two different views of the target.
By employing this method, we call the resulting representations view-invariant representations. 
In theory, models trained with these representations are better at generalizing across diverse viewing conditions, making them more robust in real-world scenarios with frequent viewpoint changes.
View-invariant representations are useful to trackers for effective tracking since extreme changes in viewing angles are a common challenge in UAV tracking.

To strengthen the single-stream paradigm for real-time UAV tracking, we conduct extensive experiments, highlight the two core design principles of AVTrack , and introduce an improved version called AVTrack-MD.
Specifically, we introduce a novel multi-teacher knowledge distillation (MD) framework based on MI maximization into the AVTrack. 
In the proposed MD, three off-the-shelf tracking models from the AVTrack serve as teachers, distilling knowledge into a more lightweight student model. 
The student model adopts the same structure as the teacher models but uses a smaller ViT backbone, containing only half the number of blocks.
In practice, multiple teacher models offer comprehensive guidance by providing diverse knowledge, which is advantageous for training the student \cite{gou2021knowledge,wang2021knowledge}.
Multi-teacher knowledge distillation is an effective strategy that is widely used across various computer vision tasks, including image super-resolution \cite{jiang2024mtkd}, image classification \cite{wen2024class}, and visual retrieval \cite{ma2024let}.
However, to the best of our knowledge, its effectiveness in UAV tracking has not been extensively investigated.
In our work, given that the mean squared error (MSE) is sensitive to noise and outliers \cite{hinton2015distilling}, we maximize the MI between the aggregated softened features of the multi-teacher models and the student model's softened features, leading to improved generalization and performance of the student model, particularly in noisy conditions.
Extensive experimental results demonstrate that the proposed method can produce a high performing student network, with overall tracking performance that is comparable to or even exceeds that of teacher networks, while using significantly fewer parameters and achieving faster speeds (see Fig. \ref{fig_Prec_Speed} and Table \ref{table_lightweight_trackers}).

This work is an extension of our previous conference paper, AVTrack \cite{lilearning2024}, accepted at ICML 2024.
In this extended paper, we propose AVTrack-MD, an improved tracker that significantly enhances the efficiency of AVTrack while achieving the overall tracking performance comparable to or even superior to it.
We also provide extensive experiments and detailed implementations.
The main contributions of our work are summarized as follows:

\begin{enumerate}

\item We propose AVTrack, an efficient real-time tracking framework that dynamically activates transformer blocks and learns view-invariant representations by maximizing mutual information between different target views, achieving promising performance while maintaining extremely fast tracking speeds. 

\item We further propose AVTrack-MD, a lightweight variant of AVTrack designed specifically for UAV tracking.
It incorporates simplified architectural designs that significantly improve computational efficiency while maintaining or even surpassing the accuracy of the original model.

\item We have proposed additional model variants and conducted experiments on more UAV tracking benchmarks, such as WebUAV-3M~\cite{zhang2022webuav}, UAVTrack112L~\cite{fu2021onboard}, and UAV20L~\cite{mueller2016benchmark} , offering extensive comparisons with recent state-of-the-art trackers along with comprehensive evaluations and analyses.

\item We also provide more detailed discussions on design principles and implementation strategies to clearly illustrate our approach and serve as a reference for designing more effective and efficient trackers.

\end{enumerate}

\section{RELATED WORKS}\label{section_related_work}


\subsection{UAV Tracking}
There are two main types of modern UAV trackers: DCF-based~\cite{lin2020learning,danelljan2015learning,li2021learning,li2020autotrack,Huang2019LearningAR,kiani2017learning,dai2019visual} and DL-based trackers \cite{fu2021siamese,liu2023bactrack,yuan2024multi,cao2021hift,cao2022tctrack}.
For instance, TB-BiCF~\cite{lin2020learning} introduces a bidirectional incongruity-aware correlation filter that leverages a temporary block to capture and store inter-frame information for short durations.
MT-Track~\cite{yuan2024multi} presents an optimized multi-step temporal modeling framework that effectively utilizes temporal context from past frames to enhance UAV tracking.
DCF-based trackers offer efficient performance but lack robustness due to limited feature representation, while DL-based trackers are constrained by high computational costs and sluggish performance, limiting their practical applicability.
Although model compression techniques, as seen in \cite{wang2022rank}, were utilized to enhance efficiency, these trackers still face challenges associated with unsatisfactory tracking precision.
A recent trend in the visual tracking community shows an increasing preference for single-stream architectures that seamlessly integrate feature extraction and correlation using pre-trained ViT backbone networks \cite{Xie2021LearningTR, cui2022mixformer,ye2022joint, Xie2022CorrelationAwareDT,xie2024autoregressive,kang2023exploring,wei2023autoregressive}, demonstrating significant potential.
Several representative methods, including OSTrack \cite{ye2022joint}, Mixformer \cite{cui2022mixformer}, DropMAE \cite{wu2023dropmae}, ZoomTrack \cite{kou2024zoomtrack}, AQATrack \cite{xie2024autoregressive}, EVPTrack \cite{shi2024evptrack}, etc., demonstrate the significant success of applying this paradigm to tracking tasks.
Although these frameworks are efficient due to their compact nature, very few are based on lightweight ViTs, making them impractical for real-time UAV tracking.
To address this, several studies have started exploring the use of efficient ViTs built on one-stream framework for real-time UAV tracking~\cite{li2023adaptive,wu2024learning,li2024learning,zhang2024efficient}.
Specifically, ETDMT \cite{zhang2024efficient} builds on lightweight ViT to introduce a tracker that combines template distinction with temporal context, while Aba-ViTrack \cite{li2023adaptive} enhances efficiency in real-time UAV tracking using lightweight ViTs and an adaptive background-aware token computation method.
However, the variable token number in \cite{li2023adaptive} necessitates unstructured access operations, leading to notable time costs.
Recent research has focused on improving the efficiency of ViTs by balancing their representation capabilities with computational efficiency. Methods like lightweight ViTs, model compression, and hybrid designs~\cite{Wang2020LinformerSW,Zhang2022MiniViTCV,Li2022EfficientFormerVT,yang2023skeleton} have been explored, but they often sacrifice accuracy or require time-consuming fine-tuning.
Recent developments in efficient ViTs with conditional computation focus on adaptive inference, dynamically adjusting computational load based on input complexity to accelerate model performance.
For example, DynamicViT~\cite{Rao2021DynamicViTEV} introduces control gates to selectively process tokens, while A-ViT~\cite{yin2022vit} employs an Adaptive Computation Time strategy to avoid auxiliary halting networks, achieving gains in efficiency, accuracy, and token prioritization.
Aba-ViTrack~\cite{li2023adaptive} effectively utilizes the latter, showcasing significant potential for real-time UAV tracking.
In our work, we focus on improving the efficiency of ViTs for UAV tracking through more structured methods, specifically the adaptive activation of transformer blocks for feature representation.

\subsection{View-Invariant Feature Representation}

View-invariant feature representation has garnered significant attention in the field of computer vision and image processing. This technique aims to extract features from images or visual data that remain consistent across various viewpoints or orientations, providing robustness to changes in the camera angle or scene configuration~\cite{kumie2024dual,Bracci2018ViewinvariantRO,Li2017DeepGaitAL,Rao2002ViewInvariantRA}.
Traditional methods typically rely on handcrafted features and geometric transformations to achieve view-invariant representations~\cite{Xia2012ViewIH,Ji2010AdvancesIV,Rao2002ViewInvariantRA}.
While effective, these traditional methods are often limited to specific scenarios with relatively simple and fixed backgrounds, making them unsuited for handling the complexity and variability of real-world visual data.
In recent years, Convolutional Neural Networks (CNNs) and other deep learning architectures have been widely adopted to extract view-invariant representations, due to their proven ability to learn highly discriminative features from complex data~\cite{kumie2024dual,men2023focalized,gao2021view,Shiraga2016GEINetVG}.
For instance, Kumie et al.~\cite{kumie2024dual} proposed the Dual-Attention Network (DANet) for view-invariant action recognition, which integrates relation-aware spatiotemporal self-attention and cross-attention modules to effectively learn representative and discriminative action features. Gao et al.~\cite{gao2021view} introduced the View Transformation Network (VTN) that realizes the view normalization by transforming arbitrary-view action samples to a base view to seek a view-invariant representation.
These approaches leverage the capacity of deep models to capture complex patterns and variations in visual data. By ensuring that learned features are resilient to changes in viewpoint, these methods contribute to the robustness and generalization of vision-based systems.
View-invariant feature representation is helpful for a variety of vision tasks, such as action recognition~\cite{Xia2012ViewIH,kumie2024dual}, pose estimation~\cite{Bracci2018ViewinvariantRO}, human re-identification \cite{liu2023learning}, and object detection~\cite{Feng2022AeDetAM}.
However, to the best of our knowledge, the integration of view-invariant representations into visual tracking frameworks remains underexplored.
In our work, we make the first attempt to learn view-invariant feature representations through mutual information maximization based on ViTs, specifically tailored for UAV tracking. This marks the first instance where ViTs are employed to acquire view-invariant feature representations in the context of UAV tracking.

\begin{figure*}[!t]
	\centering
\includegraphics[width=0.9\textwidth]{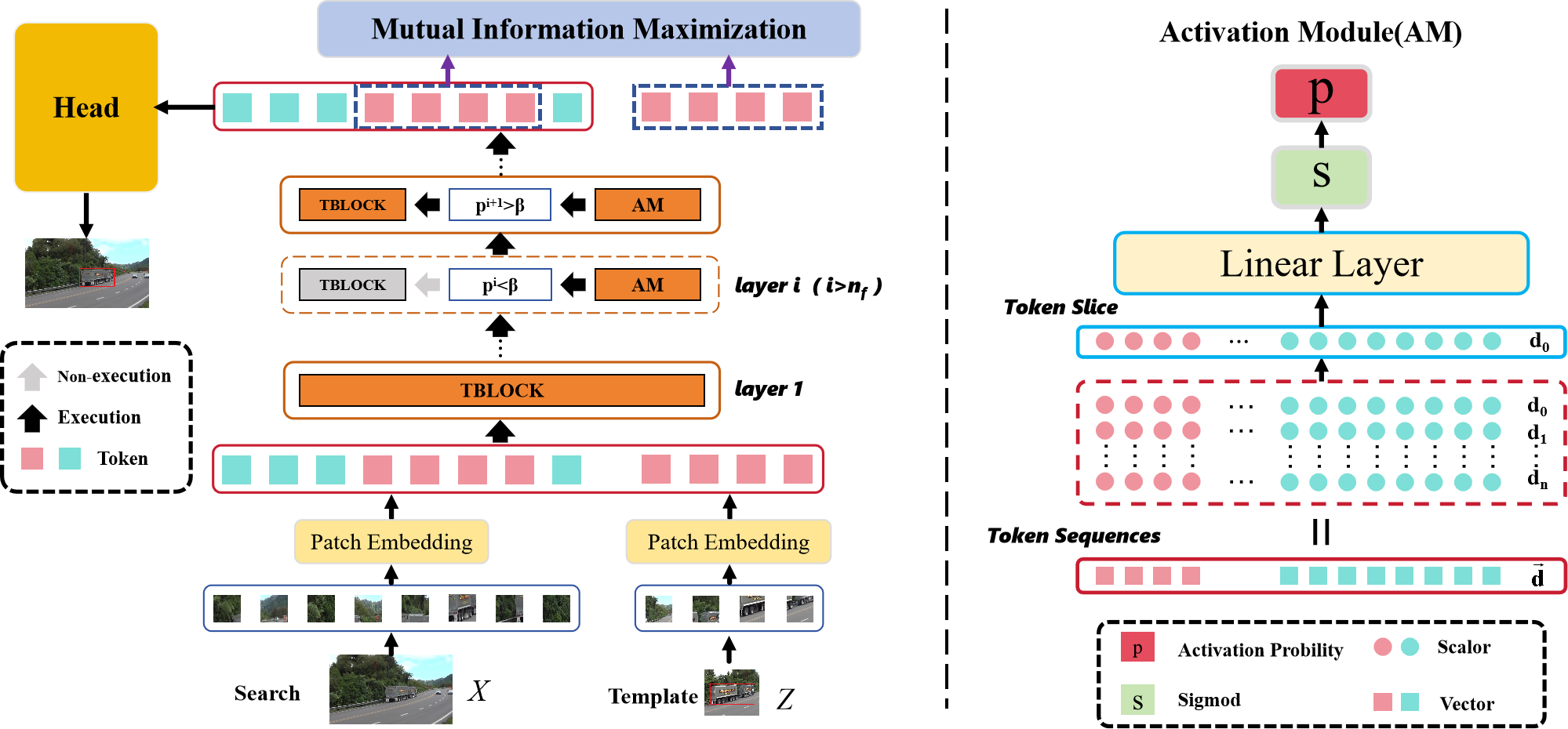}
\caption{(left) Overview of the proposed AVTrack's framework, which consists of a single-stream backbone and a prediction head. Activattion Modules (AMs) are added into transformer blocks (TBlocks) to make the ViT adaptive and Mutual Information Maximization is employed to learn View-Invariance Representations (VIR).
(right) The detailed structure of the Activation Module (AM).}
\label{fig_AVTrack_overview}
\end{figure*}

\subsection{Multi-Teacher Knowledge Distillation}

Multi-teacher knowledge distillation (MD) leverages knowledge from multiple off-the-shelf pre-trained teacher models to guide the student, enhancing its generalization by inheriting diverse knowledge from various teachers~\cite{gou2021knowledge,wang2021knowledge}.
It has primarily been studied in the contexts of image classification \cite{you2017learning,lan2024m2kd,wen2024class}, action recognition~\cite{wu2019multi} and visual retrieval \cite{ma2024let}.
For instance, MTD~\cite{wen2024class} explored a multi-teacher distillation approach for class-incremental learning, leveraging weight permutation, feature perturbation, and diversity regularization to ensure diverse mechanisms in teachers.
To better leverage the complementary knowledge of each modality, M2KD~\cite{lan2024m2kd} propose a multi-teacher multi-modal knowledge distillation framework to guide the training of the multi-modal fusion network and further improve the multi-modal feature fusion process.
Whiten-MTD \cite{ma2024let} aligns the outputs of teacher models by whitening their similarity distributions and identifies the most effective fusion strategy for their multi-teacher distillation framework through empirical analysis.
To the best of our knowledge, existing knowledge distillation methods for visual tracking rely on a single-teacher framework~\cite{li2022mask,sun2023siamohot}, limiting their ability to fully utilize the diverse knowledge from multiple teachers, potentially leading to suboptimal tracking performance.
In this work, we propose a simple yet effective MI maximization-based multi-teacher knowledge distillation framework, integrated into AVTrack, to develop a more efficient UAV tracker.
We adopt all three off-the-shelf trackers from AVTrack (i.e.,AVTrack-DeiT, AVTrack-ViT, and AVTrack-EVA) as teacher models, which offer a diverse and high-quality selection of teachers, eliminating the need for additional operations, like those in~\cite{wen2024class}, to implement varied mechanisms in the teachers.
To ensure the student model captures the most relevant information from the teacher models' representations, we propose maximizing the MI between the averaged softened feature representation of the multi-teacher models and the student model's softened feature representation.
Averaging the predictions of all teacher models is a common and effective method, as it reduces biases and provides a more objective output than any single teacher's prediction \cite{buciluǎ2006model,ba2014deep,ilichev2021multiple,song2018collaborative}.

\section{Method}\label{section_method}
In this section, we will start by introducing a brief overview of our AVTrack framework, as shown in Fig. \ref{fig_AVTrack_overview}.
Then, we detail the two proposed components: \textbf{(1)} the Activation Module (AM) for dynamically activating transformer blocks based on inputs and \textbf{(2)} the method for learning view-invariant representations (VIR) via mutual information maximization.
Additionally, we present the details of the improved tracker, AVTrack-MD, which is based on AVTrack and incorporates the MI maximization-based multi-teacher knowledge distillation framework (see Fig. \ref{fig_overview_mtkd}).
At the end of this section, we will provide a detailed introduction to the prediction head and training loss.

\begin{figure}[!h]
\centering
\includegraphics[width=0.475\textwidth]{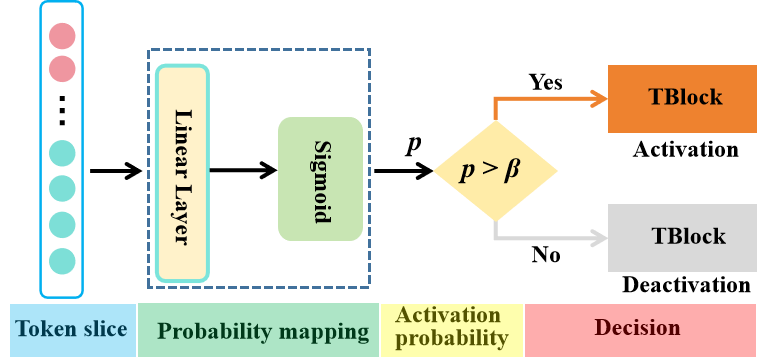}
\caption{An overview of the AM’s internal working mechanism. Note that $p$ and $\beta$ denote the activation probability and its threshold, respectively.}
\label{fig_AM_intern_work}
\end{figure}

\subsection{Overview}
The proposed AVTrack~\cite{lilearning2024} introduces a novel single-stream tracking framework, featuring an adaptive ViT-based backbone and a prediction head.
To improve the efficiency of the ViT, each transformer block in the ViT-based backbone, except for the first $n_{f}$ blocks, incorporates an Activation Module (AM). This module is trained to adaptively determine whether to activate the associated transformer block.
As shown in Fig.\ref{fig_AM_intern_work}, the internal mechanism operates as follows: AM first slices the input tokens and passes them through a Linear layer followed by a Sigmoid function to compute the activation probability. For $i$-th ViT block ($i > n_f$), if its corresponding probability exceeds a predefined threshold, the block is activated; otherwise, it is deactivated. This mechanism enables adaptive, input-dependent computation and enhances the model’s efficiency.
The framework takes a pair of images as input, comprising a template denoted as $Z\in\mathbb{R}^{3\times H_{z}\times W_{z}}$ and a search image denoted as $X\in\mathbb{R}^{3\times H_{x}\times W_{x}}$. These images are split into patches of size $P\times P$, and the number of patches for $Z$ and $X$ are $P_{z}=H_{z}\times W_{z}/P^2$ and $P_{x}=H_{x}\times W_{x}/P^2$, respectively. The features extracted from the backbone are fed into the prediction head to generate tracking results.
To obtain view-invariant representation with ViTs, we maximize the mutual information (MI) between the feature representations of two different views of the target, i.e., the template image and the target patch in the search image. During the training phase, since the ground truth localization of the target in the search image is known, we can obtain the feature representation of the subsequent view from the representation of the search image using interpolation techniques.

Building on the proposed AVTrack, we introduce a simple yet effective MI maximization-based multi-teacher knowledge distillation (MD) framework to develop a more efficient UAV tracker, called AVTrack-MD.
During the training of the AVTrack-MD model (i.e., the student model), the weights of AVTrack (i.e., the teacher model) remain fixed, while both teacher and student models receive the same inputs $Z$ and $X$.
Let $\mathfrak{B}_T$ and $\mathfrak{B}_S$ represent the backbones of the teacher and student, respectively.
In our implementation, $\mathfrak{B}_S$ shares the same structure as the $\mathfrak{B}_T$ but uses a smaller ViT layer.
Feature-based knowledge distillation transfers knowledge from the teacher models' backbone features to the student model by maximizing the MI between the averaged softened features of the multi-teacher models and the student model's softened features.
The details of these design principles will be elaborated in the subsequent subsections.

\subsection{Activation Module (AM)}
\label{AVTrack_AM}

The Activation Module selectively activates transformer blocks based on input, enabling adaptive adjustments to the ViT architecture and ensuring that only the essential blocks are activated for effective tracking. 
To enhance efficiency, the AM processes a subset of tokens representing both the target template and the search image. Its output determines whether the current transformer block is activated. If not, the tokens from the preceding block are directly propagated to the next without further processing. This mechanism is consistently applied across all transformer blocks.
Specifically, let's consider the $i$-th layer ($i > n_{f}$). 
We denote the total number of tokens by $\mathcal{K}$, the embedding dimension of each token by $d$, and all the tokens output by the  $(i-1)$-th layer by $\mathbf{t}_{1:\mathcal{K}}^{i-1}(Z,X)\in \mathbb{R}^{\mathcal{K}\times d}$. The slice of all tokens generated by the $(i-1)$-th transformer block is expressed as
$\mathbf{e}_1^{\textup{T}}\mathbf{t}_{1:\mathcal{K}}^{i-1}(Z,X):=\textbf{r}^{i-1}\in \mathbb{R}^{\mathcal{K}}$, where $\mathbf{e}_1^{\textup{T}}=[1,0,...,0]\in \mathbb{R}^{\mathcal{K}}$ is a standard unit vector in $\mathbb{R}^{\mathcal{K}}$, which is used to extract the first token slice from all tokens output by the $(i-1)$-th transformer block. This design choice serves to isolate the specific token representation $r^{i-1}$ from the full token set $\mathbf{t}_{1:\mathcal{K}}^{i-1}(Z,X)$ before further processing via the linear layer $\mathfrak{L}^i$. This token carries global contextual information and serves as a compact representation for determining the activation state of the subsequent transformer block via the AM. This design ensures minimal overhead while retaining high-level semantics for adaptive computation.
Formally, the Activation Module (AM) at layer $i$ is expressed as:
\begin{equation}
    p^i=\sigma(\mathfrak{L}^i(\textbf{r}^{i-1})),
\end{equation}
where $p^i\in [0,1]$ represents the activation probability of the $i$-th transformer block, $\sigma(x)=1/(1+e^{-x})$ indicates the sigmoid function. 
The sigmoid function is versatile and finds applications in various machine learning tasks where non-linear transformations or probabilistic outputs are required. In our work, the activation mechanism in each transformer block is determined by the sigmoid function, which helps introduce non-linearity into the AM's decision-making process. Specifically, the sigmoid function maps the output value of the linear layer to a range between 0 and 1, which represents the activation probability of the associate transformer block. 
If $p^i>\beta $, where $\beta \in (0.5, 1)$ is the activation probability threshold, the transformer block at layer $i$ will be activated; otherwise, it is deactivated and the output tokens from the $(i-1)$-th layer will be fed into the $(i+1)$-th  block directly. 
Let $\mathcal{N}$ denote the total number of transformer blocks in the given ViT. Theoretically, deactivating all $\mathcal{N}$ blocks simultaneously would result in no correlation being computed between the template and search image. To avoid such unfavorable conditions, the first $n_{f}$ layers are mandated to remain activated. This strategy helps alleviate computational burdens associated with AM, as these initial layers are typically essential, providing foundational information on which high-level and more abstract features and representations can be built.
Another extreme case occurs when all transformer blocks are activated for any input, allowing the model to more efficiently minimize classification and regression losses, as larger models have greater fitting capacity.
To address this, we introduce a block sparsity loss, $\mathcal{L}_{spar}$, which penalizes a higher mean probability across all adaptive layers, encouraging the deactivation of many blocks on average to improve efficiency. The block sparsity loss is defined as follows:
\begin{equation}
\begin{split}
    \mathcal{L}_{spar}=|
    \frac{1}{\mathcal{N}-n_{f}}\sum^{\mathcal{N}}_{i=n_{f}+1}p_i-\zeta|,
\end{split}
\end{equation}
where $\zeta \in [0,1]$ is a constant used with $\beta$ to control block sparsity. Generally, for a given $\beta$, a smaller $\zeta$ leads to a sparser model. When $\zeta = 0$, $p_i$ can be viewed as the weight of block $i$, and the sparsity loss becomes proportional to the $l_1$ norm of the vector of these weights, which is a convex-relaxed sparsity regularization commonly used in statistical learning theory. $\zeta$ serves as a hyperparameter for finer adjustments.

\subsection{ View-Invariant Representations (VIR) via Mutual Information Maximization}
\label{AVTrack_VIR}
We begin by introducing the concept of mutual information (MI) and establishing the relevant notations.
Given two random variables: $\textbf{a}\in\mathcal{A}$ and $\textbf{b}\in\mathcal{B}$.
The MI between \(\mathbf{a}\) and \(\mathbf{b}\), denoted as \(\textup{I}(\mathbf{a}, \mathbf{b})\), is expressed as follows:
\begin{equation}
\label{eq_mi}
    \mathbb{E}_{p(\textbf{a},\textbf{b})}\left[log\frac{p(\textbf{a},\textbf{b})}{p(\textbf{a})p(\textbf{b})}\right] = \textup{D}_{KL}(p(\textbf{a},\textbf{b})||p(\textbf{a})p(\textbf{b})),
\end{equation}
where $p(\textbf{a},\textbf{b})$ represents the joint probability distribution, while $p(\textbf{a})$ and $p(\textbf{b})$ are the marginal distributions. The symbol $\textup{D}_{KL}$ denotes the Kullback–Leibler divergence (KLD) \cite{MacKay2004InformationTI}. 
Estimating MI is not an easy task in real-world situations, as we usually only have access to the samples that are readily available, not the underlying distributions \cite{Poole2019OnVB}.
As a result, most existing estimators typically rely on approximating the MI between variables using observed samples.
In contrast to these approaches, we employ the Deep InfoMax MI estimator \cite{Hjelm2019LearningDR}, which estimates MI based on Jensen-Shannon divergence (JSD).
This strategy has proven to be effective, as knowing its precise value is less critical than maximizing MI in this context.
The Jensen-Shannon MI estimator, represented by $\hat{\textup{I}}_{\mathbf{\mathbf{\Theta}}}^{(JSD)}(\textbf{a},\textbf{b})$, is defined by:
\begin{equation}
    \label{eq_jsd}
    \mathbb{E}_{p(\textbf{a},\textbf{b})}[-\alpha(-T_{\mathbf{\mathbf{\Theta}}}(\textbf{a},\textbf{b}))]-\mathbb{E}_{p(\textbf{a})p(\textbf{b})}[\alpha(T_{\mathbf{\mathbf{\Theta}}}(\textbf{a},\textbf{b}))],
\end{equation}
where $T_{\mathbf{\mathbf{\mathbf{\Theta}}}}: \mathit{X} \times \mathit{Y}\rightarrow \mathbb{R}$ is a neural network parameterized by $\mathbf{\mathbf{\mathbf{\Theta}}}$, and $\alpha(z)=log(1+e^z)$ represents the softplus function.

In our work, the proposed approach involves learning view-invariant feature representations by maximizing MI using the aforementioned Jensen-Shannon MI estimator between the feature representations of two different views of a specific target.
Let $\mathbf{t}_{1:\mathcal{K}}^{\infty}(Z,X)=\mathbf{t}_{\mathcal{K}_Z\cup \mathcal{K}_X}^{\infty}(Z,X)$,  $\mathcal{K}_Z
\cup\mathcal{K}_X=[1,\mathcal{K}]$,
denote the final output tokens of the ViT, where $\mathbf{t}_{\mathcal{K}_Z}^{\infty}$ and $\mathbf{t}_{\mathcal{K}_X}^{\infty}$ represent the tokens corresponding to the template and the search image, respectively. 
Given the ground truth localization of the target, denoted as $Z'$ in the search image, we obtain the tokens corresponding to $Z'$ through linear interpolation, represented by $\mathbf{t}_{\mathcal{K}_{Z'}}^{\infty}(Z,X) \subset \mathbf{t}_{\mathcal{K}_X}^{\infty}(Z,X)$.
The specially designed loss \(\mathcal{L}_{vir}\) for learning view-invariant feature representations is formulated as follow,
\begin{equation}
    \label{eq_mi}
    \mathcal{L}_{vir} = -\hat{\textup{I}}^{(JSD)}_{\mathbf{\mathbf{\mathbf{\Theta}}}}(\mathbf{t}_{\mathcal{K}_{Z'}}^{\infty}(Z,X),\mathbf{t}_{\mathcal{K}_{Z}}^{\infty}(Z,X)).
\end{equation}
During the inference phase, only the sequence \([Z, X]\) is input into the VIR, and the process for learning view-invariant representations is not involved.
Consequently, our method imposes no additional computational cost during the inference phase.
Additionally, the proposed view-invariant representation learning is ViT-agnostic, making it easily adaptable to other tracking frameworks.

\begin{figure}[!h]
\centering
\includegraphics[width=0.475\textwidth]{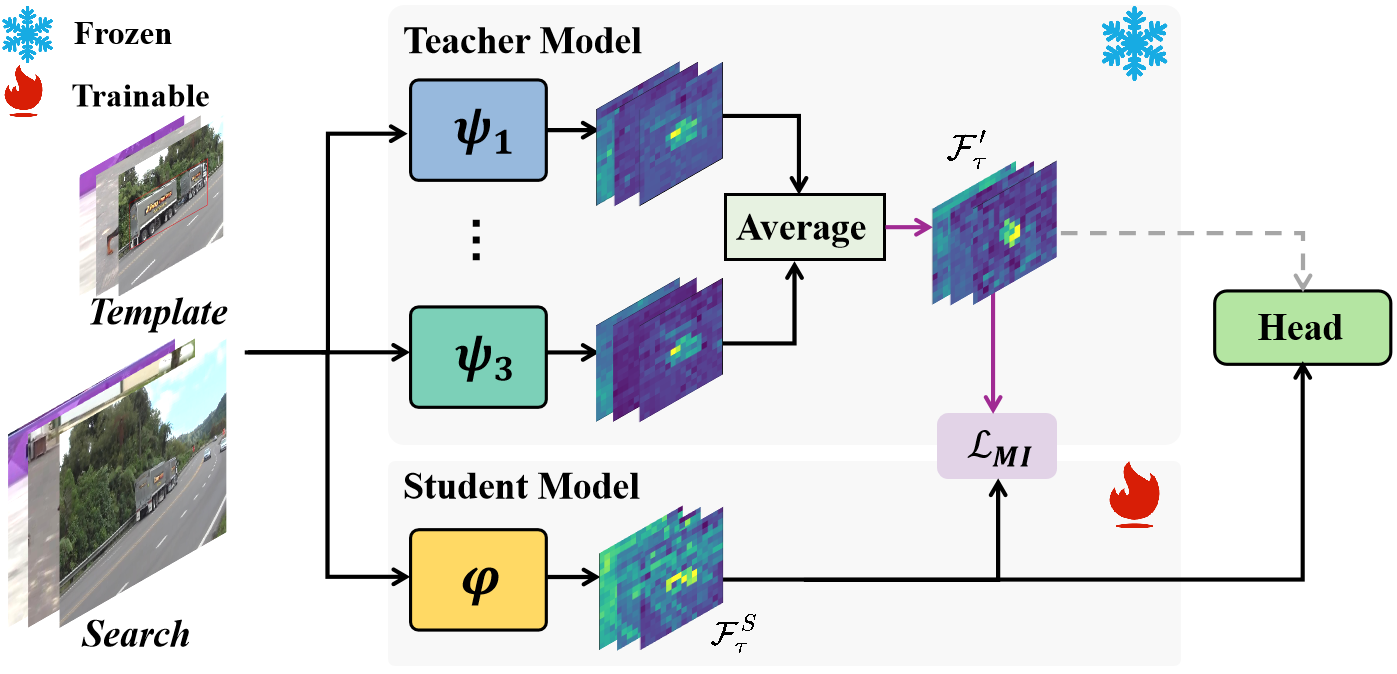}
\caption{An overview of the proposed MI maximization-based multi-teacher knowledge distillation. Note that $\mathcal{F}_\tau’$ denotes the randomly shuffled of $\mathcal{F}_\tau $.}
\label{fig_overview_mtkd}
\end{figure}

\subsection{MI Maximization-based Multi-Teacher Knowledge Distillation (MD)}

We developed AVTrack-MD with the aim of achieving performance comparable to or superior to the AVTrack while reducing computational resources and memory usage, making it a more efficient UAV tracker.
To achieve this, we introduce a novel MI maximization-based multi-teacher knowledge distillation framework into the AVTrack, as illustrated in Fig. \ref{fig_overview_mtkd}.
Our method focuses on feature-level knowledge distillation, meaning that the student model learns from an ensemble of teacher models by concentrating on the features the teacher models have learned.

In the context of multi-teacher knowledge distillation, the selection of teachers-student architectures plays a crucial role in the knowledge distillation process, as it determines how the knowledge from the various teacher models is transferred to the student model.
For the teacher models, we use all three off-the-shelf trackers from AVTrack (i.e., AVTrack-DeiT, AVTrack-ViT, and AVTrack-EVA), providing diverse and high-quality teachers without the need for additional operations to implement varied mechanisms in teachers.
For the selection of student models, we opt for a self-similar architecture, where the student shares the same structure as the teacher but uses a smaller ViT backbone (i.e., with half ViT blocks). 
This selection offers two main advantages: 1) it eliminates the need for complex design processes, enabling straightforward implementation and training, and enhancing the interpretability of the model's behavior; 2) it promotes modularity and scalability, reducing the difficulty of expanding or modifying the model as needed.
Averaging the predictions of all teacher models is a common choice \cite{buciluǎ2006model,ba2014deep,ilichev2021multiple,song2018collaborative}, as it is a straightforward and effective method. 
In our implementation, we average the $K$ feature representations $\mathcal{F}_k$ produced by the teacher models into an aggregated feature representation $\mathcal{F} = \frac{1}{K} \sum_{k=1}^{K} \mathcal{F}_k.$
In practice, to provide richer information during training, the model outputs are softened using a temperature $\tau$, since the original outputs are typically represented in a one-hot encoding format \cite{you2017learning}.
Let $\mathcal{F}^S$ represent the feature representation of the student.
The softened outputs of the student and teacher are expressed as follows:
\begin{align}
\mathcal{F}_\tau = softmax(\frac{\mathcal{F}}{\tau}), 
\mathcal{F}^S_\tau = softmax(\frac{\mathcal{F}^S}{\tau}), 
\end{align}
where $\tau$ is a constant that we set to 2.
Encouraging the student network to replicate the feature representations of the teacher's final layer network using mean squared error (MSE) loss is straightforward \cite{ba2014deep,kim2021comparing,hamidi2024train}.
However, since MSE is sensitive to noise and outliers \cite{hinton2015distilling}, we maximize the MI between the aggregated softened feature of the teacher models and the student model's softened feature, enhancing generalization and performance, especially in noisy conditions.
Given the teachers-student architecture, we apply the Jensen-Shannon MI estimator to achieve MI maximization for multi-teacher knowledge distillation, which establishes the objective function for our MI maximization-based multi-teacher knowledge distillation approach, as expressed in the following equation:
\begin{align}
\label{loss_mtkd}
\mathcal{L}_{MD} = -\hat{\textup{I}}^{(JSD)}_{\mathbf{\mathbf{\mathbf{\Theta}}}}(\mathcal{F}_\tau,\mathcal{F}^S_\tau),
\end{align}
In distillation training, the student model is trained using a weighted sum of $\mathcal{L}_{MD}$ and the total loss function used for training the teacher model.

\subsection{Prediction Head and Training Objective}
Following the corner detection head in \cite{cui2022mixformer, ye2022joint}, we utilize a prediction head \( \mathcal{H} \) consisting of several Conv-BN-ReLU layers to directly estimate the target's bounding box.
The output tokens associated with the search image are first reinterpreted into a 2D spatial feature map, which is then input to the prediction head. The head generates three outputs: a target classification score $\mathbf{p} \in [0,1]^{H_x/P\times W_x/P}$, a local offset $\mathbf{o}\in [0,1]^{2\times H_x/P\times W_x/P}$, and a normalized bounding box size $\mathbf{s} \in [0,1]^{2\times H_x/P\times W_x/P}$.
The highest classification score is used to estimate the coarse target position, i.e., $(x_c, y_c)=\textup{argmax}_{(x,y)}\mathbf{p}(x,y)$, based on which the final target bounding box is determined by:
\begin{equation}
		\{(x_t,y_t);(w,h)\}=\{(x_c, y_c)+\mathbf{o}(x_c,y_c);\mathbf{s}(x_c,y_c)\}.
\end{equation}
For the tracking task, we employ the weighted focal loss \cite{Law2018CornerNetDO} for classification and a combination of $L_1$ loss and GIoU loss \cite{Rezatofighi2019GeneralizedIO} for bounding box regression.
The total loss function for tracking prediction is defined as follows:
\begin{equation}
\label{eq4}
 \mathcal{L}_{pred} =  \mathcal{L}_{cls} + \lambda_{iou} \mathcal{L}_{iou} + \lambda_{L_1} \mathcal{L}_{L1}
\end{equation}
where the constants $\lambda_{iou}=2$ and $\lambda_{L_1}=5$ are the same as in \cite{ye2022joint}.
The AVTrack framework is trained end-to-end using the overall loss $\mathcal{L}_{overall}=\mathcal{L}_{pred}+ \gamma \mathcal{L}_{spar} + \kappa \mathcal{L}_{vir}$, where $\gamma$ is set to 50 and $\kappa$ to 0.0001 after initialization with pretrained ViT weights.
After this training, we apply the proposed MI maximization-based multi-teacher knowledge distillation framework to obtain a student model that better balances performance and efficiency. Specifically, during the distillation phase, the total loss $\mathcal{L}_{overall}^*=\mathcal{L}_{pred}+\eta\mathcal{L}_{MD}$ is employed for end-to-end distillation training, with the weight $\eta$ of $\mathcal{L}_{MD}$ is set to $0.1 \times 10^{-3}$.

\begin{table*}[t]
\scriptsize
\centering
\setlength\tabcolsep{5.5pt}
\caption{Comparison of precision (Prec.), success rate (Succ.), and speed (FPS) between our trackers and lightweight trackers on DTB70 \cite{li2017visual}, UAVDT \cite{du2018the}, VisDrone2018 \cite{wen2018visdrone}, UAV123 \cite{mueller2016benchmark}, and UAV123@10fps \cite{mueller2016benchmark}. \textcolor{red}{ \textbf{Red}}, \textcolor{blue}{ \textbf{blue}}, and {\color[HTML]{1f821c}\textbf{green}} signify the first, second, and third places. Please note that the percent symbol (\%) is excluded for Prec. and Succ. values.}
\label{table_lightweight_trackers}
\begin{tabular}{ccccccccccccccccc}
\toprule[1pt]
\multicolumn{2}{c}{}                                                                   &                                                  & \multicolumn{2}{c}{DTB70}                                                                                                   & \multicolumn{2}{c}{UAVDT}                                                                                                   & \multicolumn{2}{c}{VisDrone2018}                                                                                                & \multicolumn{2}{c}{UAV123}                                                                                                  & \multicolumn{2}{c}{UAV123@10fps}                                                                                            & \multicolumn{2}{c}{Avg.}                                                                                                    & \multicolumn{2}{c}{Avg.FPS}                                                                           \\
\multicolumn{2}{c}{\multirow{-2}{*}{Method}}                                           & \multirow{-2}{*}{Source}                         & Prec.                                                        & Succ.                                                        & Prec.                                                        & Succ.                                                        & Prec.                                                        & Succ.                                                        & Prec.                                                        & Succ.                                                        & Prec.                                                        & Succ.                                                        & Prec.                                                        & Succ.                                                        & GPU                                                           & CPU                                   \\ \hline
                             & KCF\cite{henriques2015high}                                                     & TAPMI 15                                         & 46.8                                                         & 28.0                                                         & 57.1                                                         & 29.0                                                         & 68.5                                                         & 41.3                                                         & 52.3                                                         & 33.1                                                         & 40.6                                                         & 26.5                                                         & 53.1                                                         & 31.6                                                         & -                                                             & {\color[HTML]{FE0000} \textbf{622.5}} \\
                             & fDSST\cite{danelljan2017discriminative}                                                   & TPAMI 17                                         & 53.4                                                         & 35.7                                                         & 66.6                                                         & 38.3                                                         & 69.8                                                         & 51.0                                                         & 58.3                                                         & 40.5                                                         & 51.6                                                         & 37.9                                                         & 60.0                                                         & 40.7                                                         & -                                                             & {\color[HTML]{3531FF} \textbf{193.4}} \\
                             & ECO\_HC\cite{danelljan2017eco}                                                 & CVPR 17                                          & 63.5                                                         & 44.8                                                         & 69.4                                                         & 41.6                                                         & 80.8                                                         & 58.1                                                         & 71.0                                                         & 49.6                                                         & 64.0                                                         & 46.8                                                         & 69.7                                                         & 48.2                                                         & -                                                             & {\color[HTML]{009901} \textbf{83.5}}  \\
                             & MCCT\_H\cite{wang2018multi}                                                 & CVPR 18                                          & 60.4                                                         & 40.5                                                         & 66.8                                                         & 40.2                                                         & 80.3                                                         & 56.7                                                         & 65.9                                                         & 45.7                                                         & 59.6                                                         & 43.4                                                         & 66.6                                                         & 45.3                                                         & -                                                             & 63.4                                  \\
                             & STRCF\cite{li2018learning}                                                   & CVPR 18                                          & 64.9                                                         & 43.7                                                         & 62.9                                                         & 41.1                                                         & 77.8                                                         & 56.7                                                         & 68.1                                                         & 48.1                                                         & 62.7                                                         & 45.7                                                         & 67.3                                                         & 47.1                                                         & -                                                             & 28.4                                  \\
                             & ARCF\cite{Huang2019LearningAR}                                                    & ICCV 19                                          & 69.4                                                         & 47.2                                                         & 72.0                                                         & 45.8                                                         & 79.7                                                         & 58.4                                                         & 67.1                                                         & 46.8                                                         & 66.6                                                         & 47.3                                                         & 71.0                                                         & 47.1                                                         & -                                                             & 34.2                                  \\
                             & AutoTrack\cite{li2020autotrack}                                               & CVPR 20                                          & 71.6                                                         & 47.8                                                         & 71.8                                                         & 45.0                                                         & 78.8                                                         & 57.3                                                         & 68.9                                                         & 47.2                                                         & 67.1                                                         & 47.7                                                         & 71.6                                                         & 49.0                                                         & -                                                             & 57.8                                  \\
\multirow{-8}{*}{\rotatebox{90}{DCF-based}}  & RACF\cite{li2022learning}                                                    & PR 22                                            & 72.6                                                         & 50.5                                                         & 77.3                                                         & 49.4                                                         & 83.4                                                         & 60.0                                                         & 70.2                                                         & 47.7                                                         & 69.4                                                         & 48.6                                                         & 74.6                                                         & 51.2                                                         & -                                                             & 35.6                                  \\ \hline
                             & HiFT\cite{cao2021hift}                                                    & ICCV 21                                          & 80.2                                                         & 59.4                                                         & 65.2                                                         & 47.5                                                         & 71.9                                                         & 52.6                                                         & 78.7                                                         & 59.0                                                         & 74.9                                                         & 57.0                                                         & 74.2                                                         & 55.1                                                         & 160.3                                                         & -                                     \\
                             & SiamAPN\cite{fu2021siamese}                                                 & ICRA 21                                          & 78.4                                                         & 58.5                                                         & 71.1                                                         & 51.7                                                         & 81.5                                                         & 58.5                                                         & 76.5                                                         & 57.5                                                         & 75.2                                                         & 56.6                                                         & 76.7                                                         & 56.6                                                         & 194.4                                                         & -                                     \\
                             & P-SiamFC++\cite{wang2022rank}                                              & ICME 22                                          & 80.3                                                         & 60.4                                                         & 80.7                                                         & 56.6                                                         & 80.1                                                         & 58.5                                                         & 74.5                                                         & 48.9                                                         & 73.1                                                         & 54.9                                                         & 77.7                                                         & 55.9                                                         & 240.5                                                         & 56.9                                  \\
                             & TCTrack\cite{cao2022tctrack}                                                 & CVPR 22                                          & 81.1                                                         & 61.9                                                         & 69.1                                                         & 50.4                                                         & 77.6                                                         & 57.7                                                         & 77.3                                                         & 60.4                                                         & 75.1                                                         & 58.8                                                         & 76.0                                                         & 57.8                                                         & 139.6                                                         & -                                     \\
                             & TCTrack++\cite{cao2023towards}                                               & TPAMI 23                                         & 80.4                                                         & 61.7                                                         & 72.5                                                         & 53.2                                                         & 80.8                                                         & 60.3                                                         & 74.4                                                         & 58.8                                                         & 78.2                                                         & 60.1                                                         & 77.3                                                         & 58.8                                                         & 125.6                                                         & -                                     \\
                             & SGDViT\cite{yao2023sgdvit}                                                  & ICRA 23                                          & 78.5                                                         & 60.4                                                         & 65.7                                                         & 48.0                                                         & 72.1                                                         & 52.1                                                         & 75.4                                                         & 57.5                                                         & {\color[HTML]{FE0000} \textbf{86.3}}                         & {\color[HTML]{FE0000} \textbf{66.1}}                         & 75.6                                                         & 56.8                                                         & 110.5                                                         & -                                     \\
                             & ABDNet\cite{zuo2023adversarial}                                                  & RAL 23                                           & 76.8                                                         & 59.6                                                         & 75.5                                                         & 55.3                                                         & 75.0                                                         & 57.2                                                         & 79.3                                                         & 60.7                                                         & 77.3                                                         & 59.1                                                         & 76.7                                                         & 59.1                                                         & 130.2                                                         & -                                     \\
& DRCI\cite{zeng2023towards}                                                    & ICME 23                                          & 81.4                                                         & 61.8                                                         & {\color[HTML]{FE0000} \textbf{84.0}}                         & 59.0                                                         & 83.4                                                         & 60.0                                                         & 76.7                                                         & 59.7                                                         & 73.6                                                         & 55.2                                                         & 79.8                                                         & 59.1                                                         & 281.3                                                         & 62.4                                  \\
\multirow{-9}{*}{\rotatebox{90}{CNN-based}}  & PRL-Track \cite{fu2024progressive}                                            & IROS 24                                                 & 79.5                                                         & 60.6                                                         & 73.1                         & 53.5                                                         & 72.6                                                        & 53.8                                                        & 79.1                                                         & 59.3                                                        &74.1                                                         & 58.6
&75.2
&57.2
& 135.6                                                         & -                                  \\ \hline
                             & Aba-ViTrack\cite{li2023adaptive}                                             & ICCV 23                                          & {\color[HTML]{FE0000} \textbf{85.9}}                         & {\color[HTML]{FE0000} \textbf{66.4}}                         & {\color[HTML]{3531FF} \textbf{83.4}}                         & {\color[HTML]{009901} \textbf{59.9}}                         & {\color[HTML]{3531FF} \textbf{86.1}}                         & {\color[HTML]{3531FF} \textbf{65.3}}                         & {\color[HTML]{FE0000} \textbf{86.4}}                         & {\color[HTML]{009901} \textbf{66.4}}                         & {\color[HTML]{3531FF} \textbf{85.0}}                         & 65.5                                                         & {\color[HTML]{FE0000} \textbf{85.3}}                         & {\color[HTML]{FE0000} \textbf{64.7}}                         & 181.5                                                         & 50.3                                  \\
                             & LiteTrack\cite{wei2024litetrack}                                               & ICRA 24                                          & 82.5                                                         & 63.9                                                         & 81.6                                                         & 59.3                                                         & 79.7                                                         & 61.4                                                         & 84.2                                                         & 65.9                                                         & 83.1                                                         & 64.0                                                         & 82.2                                                         & 62.9                                                         & 119.7                                                         & -                                     \\
                             & SMAT\cite{gopal2024separable}                                                    & WACV 24                                          & 81.9                                                         & 63.8                                                         & 80.8                                                         & 58.7                                                         & 82.5                                                         & 63.4                                                         & 81.8                                                         & 64.6                                                         & 80.4                                                         & 63.5                                                         & 81.5                                                         & 62.8                                                         & 124.2                                                         & -                                     \\
                             & LightFC\cite{li2024lightweight}                                                & KBS 24                                           & 82.8                                                         & 64.0                                                         & {\color[HTML]{3531FF} \textbf{83.4}}                         & {\color[HTML]{FE0000} \textbf{60.6}}                         & 82.7                                                         & 62.8                                                         & 84.2                                                         & 65.5                                                         & 81.3                                                         & 63.7                                                         & 82.9                                                         & 63.4                                                         & 146.8                                                         & -                                     \\
                             & SuperSBT\cite{xie2024correlation}                                                & TPAMI 24                                         & {\color[HTML]{009901} \textbf{84.5}}                         & {\color[HTML]{009901} \textbf{65.4}}                         & 81.5                                                         & 60.3                                                         & 80.4                                                         & 62.2                                                         & {\color[HTML]{3531FF} \textbf{85.0}}                         & {\color[HTML]{FE0000} \textbf{67.2}}                         & 82.1                                                         & 65.3                                                         & 82.7                                                         & 64.1                                                         & 121.7                                                         & -                                     \\
                             & \cellcolor[HTML]{D9D9D9}\cellcolor[HTML]{D9D9D9}\textbf{AVTrack-ViT}                                & \cellcolor[HTML]{D9D9D9}    & \cellcolor[HTML]{D9D9D9}81.3                                 & \cellcolor[HTML]{D9D9D9}63.3                                 & \cellcolor[HTML]{D9D9D9}79.9                                 & \cellcolor[HTML]{D9D9D9}57.7                                 & \cellcolor[HTML]{D9D9D9}{\color[HTML]{FE0000} \textbf{86.4}} & \cellcolor[HTML]{D9D9D9}{\color[HTML]{FE0000} \textbf{65.9}} & \cellcolor[HTML]{D9D9D9}84.0                                 & \cellcolor[HTML]{D9D9D9}66.2                                 & \cellcolor[HTML]{D9D9D9}83.2                                 &\cellcolor[HTML]{D9D9D9}65.7        & \cellcolor[HTML]{D9D9D9}82.9                                 & \cellcolor[HTML]{D9D9D9}63.1                                 & \cellcolor[HTML]{D9D9D9}250.2                                 & \cellcolor[HTML]{D9D9D9}58.7          \\
                             & \cellcolor[HTML]{D9D9D9}\textbf{AVTrack-EVA}                                & \cellcolor[HTML]{D9D9D9}     & \cellcolor[HTML]{D9D9D9}82.6                                 & \cellcolor[HTML]{D9D9D9}64.0                                 & \cellcolor[HTML]{D9D9D9}78.8                                 & \cellcolor[HTML]{D9D9D9}57.2                                 & \cellcolor[HTML]{D9D9D9}84.4                                 & \cellcolor[HTML]{D9D9D9}63.5                                 & \cellcolor[HTML]{D9D9D9}83.0                                 & \cellcolor[HTML]{D9D9D9}64.7                                 & \cellcolor[HTML]{D9D9D9}81.2                                 & \cellcolor[HTML]{D9D9D9}63.5                                 & \cellcolor[HTML]{D9D9D9}82.0                                 & \cellcolor[HTML]{D9D9D9}62.6                                 & \cellcolor[HTML]{D9D9D9}283.7                                 & \cellcolor[HTML]{D9D9D9}62.8          \\
                             & \cellcolor[HTML]{D9D9D9}\textbf{AVTrack-DeiT} & \multirow{-3}{*}{\cellcolor[HTML]{D9D9D9}\textbf{Ours}}   & \cellcolor[HTML]{D9D9D9}84.3                                 & \cellcolor[HTML]{D9D9D9}65.0                                 & \cellcolor[HTML]{D9D9D9}82.1                                 & \cellcolor[HTML]{D9D9D9}58.7                                 & \cellcolor[HTML]{D9D9D9}{\color[HTML]{009901} \textbf{86.0}} & \cellcolor[HTML]{D9D9D9}{\color[HTML]{3531FF} \textbf{65.3}} & \cellcolor[HTML]{D9D9D9}{\color[HTML]{009901} \textbf{84.8}} & \cellcolor[HTML]{D9D9D9}{\color[HTML]{3531FF} \textbf{66.8}} & \cellcolor[HTML]{D9D9D9}83.2                                 & \cellcolor[HTML]{D9D9D9}{\color[HTML]{009901} \textbf{65.8}} & \cellcolor[HTML]{D9D9D9}{\color[HTML]{3531FF} \textbf{84.1}} & \cellcolor[HTML]{D9D9D9}{\color[HTML]{3531FF} \textbf{64.4}} & \cellcolor[HTML]{D9D9D9}256.8                                 & \cellcolor[HTML]{D9D9D9}59.5  \\       
                             & \cellcolor[HTML]{eff7ff}\textbf{AVTrack-MD-ViT} & {\cellcolor[HTML]{eff7ff}}   & \cellcolor[HTML]{eff7ff}{\color[HTML]{3531FF} \textbf{84.9}}                                 & \cellcolor[HTML]{eff7ff}{\color[HTML]{3531FF} \textbf{65.7}}                                 & \cellcolor[HTML]{eff7ff}81.4                                 & \cellcolor[HTML]{eff7ff}59.5                                 & \cellcolor[HTML]{eff7ff}84.8 & \cellcolor[HTML]{eff7ff}63.7 & \cellcolor[HTML]{eff7ff}82.3 & \cellcolor[HTML]{eff7ff}65.1 & \cellcolor[HTML]{eff7ff}{\color[HTML]{009901} \textbf{83.5}}                                 & \cellcolor[HTML]{eff7ff}{\color[HTML]{3531FF} \textbf{65.9}} & \cellcolor[HTML]{eff7ff}{\color[HTML]{3531FF} \textbf{83.4}} & \cellcolor[HTML]{eff7ff}{\color[HTML]{3531FF} \textbf{64.0}} & \cellcolor[HTML]{eff7ff}{\color[HTML]{009901} \textbf{303.1}}                                & \cellcolor[HTML]{eff7ff}63.8
                             \\
                             & \cellcolor[HTML]{eff7ff} \textbf{AVTrack-MD-EVA}                                 & \cellcolor[HTML]{eff7ff}  & \cellcolor[HTML]{eff7ff}83.2                                 & \cellcolor[HTML]{eff7ff}63.9                                 & \cellcolor[HTML]{eff7ff}80.8                                 & \cellcolor[HTML]{eff7ff}58.0                                 & \cellcolor[HTML]{eff7ff}84.0                                 & \cellcolor[HTML]{eff7ff}63.5                                 & \cellcolor[HTML]{eff7ff}81.5                                 & \cellcolor[HTML]{eff7ff}62.3                                 & \cellcolor[HTML]{eff7ff}82.7                                 & \cellcolor[HTML]{eff7ff}64.7                                 & \cellcolor[HTML]{eff7ff}82.4                                 & \cellcolor[HTML]{eff7ff}62.5                                 & \cellcolor[HTML]{eff7ff}{\color[HTML]{FE0000} \textbf{334.4}} & \cellcolor[HTML]{eff7ff}67.1          \\
\multirow{-11}{*}{\rotatebox{90}{ViT-based}} & \cellcolor[HTML]{eff7ff} \textbf{AVTrack-MD-DeiT} & \multirow{-3}{*}{\cellcolor[HTML]{eff7ff}\textbf{Ours}}  & \cellcolor[HTML]{eff7ff}84.0                                 & \cellcolor[HTML]{eff7ff}65.2                                 & \cellcolor[HTML]{eff7ff}{\color[HTML]{009901}\textbf{83.1}} & \cellcolor[HTML]{eff7ff}{\color[HTML]{3531FF}\textbf{60.3}} & \cellcolor[HTML]{eff7ff}84.9                                 & \cellcolor[HTML]{eff7ff}{\color[HTML]{009901}\textbf{64.2}} & \cellcolor[HTML]{eff7ff}82.6                                 & \cellcolor[HTML]{eff7ff}65.2                                 & \cellcolor[HTML]{eff7ff}83.3                                 & \cellcolor[HTML]{eff7ff}65.5                                 & \cellcolor[HTML]{eff7ff}{\color[HTML]{009901} \textbf{83.6}} & \cellcolor[HTML]{eff7ff}{\color[HTML]{009901}\textbf{64.1}} & \cellcolor[HTML]{eff7ff}{\color[HTML]{3531FF}\textbf{310.6}} & \cellcolor[HTML]{eff7ff}64.8   \\ \bottomrule[1pt]      
\end{tabular}
\end{table*}

\section{Experiments}
\label{section_experiment}
In this section, we evaluate our tracker on six publicly widely-used UAV tracking benchmarks, including DTB70 \cite{li2017visual}, UAVDT \cite{du2018the}, VisDrone2018 \cite{wen2018visdrone}, UAV123 \cite{mueller2016benchmark}, UAV123@10fps \cite{mueller2016benchmark}, and WebUAV-3M \cite{zhang2022webuav}.
Specifically, we conduct a thorough comparison with over 38 existing state-of-the-art (SOTA) trackers, using their results obtained from running the official codes with the corresponding hyperparameters.
Please note that our evaluation is performed on a PC that was equipped with an i9-10850K processor (3.6GHz), 16GB of RAM, and an NVIDIA TitanX GPU.
To facilitate clearer comparison among different tracker types, we categorize them into two groups: lightweight trackers and deep trackers.

\subsection{Implementation Details}

\textbf{Model.}
The AVTrack~\cite{lilearning2024} features three trackers for evaluation: AVTrack-ViT, AVTrack-DeiT, and AVTrack-EVA, with each employing a lightweight ViT as its backbone: ViT-tiny~\cite{dosovitskiy2020image}, DeiT-tiny~\cite{touvron2021training}, and EVA-tiny~\cite{fang2023eva}, respectively.
Building on it, AVTrack-MD introduces the novel MI maximization-based multi-teacher knowledge distillation (MD) framework to develop three corresponding improved trackers: AVTrack-MD-ViT, AVTrack-MD-DeiT, and AVTrack-MD-EVA.
The head and input sizes for AVTrack and AVTrack-MD are configured identically. The head consists of a stack of four Conv-BN-ReLU layers, with input sizes set to $256^2$ for the search image and $128^2$ for the template image.

\textbf{Training. }
For training, we employ a combination of training sets from four different datasets: GOT-10k \cite{huang2019got}, LaSOT \cite{fan2019lasot}, COCO \cite{lin2014microsoft}, and TrackingNet \cite{muller2018trackingnet}.
It is noteworthy that all the trackers utilize the same training pipeline to ensure consistency and comparability.
The batch size is set to 32.
We use the AdamW optimizer to train the model, with a weight decay of $10^{-4}$ and an initial learning rate of $4 \times 10^{-5}$.
The total number of training epochs is fixed at 300, with 60,000 image pairs processed per epoch. The learning rate decreases by a factor of 10 after 240 epochs.
During the multi-teacher distillation phase, we use all three pre-trained AVTrack-* models as teacher models. The parameters of the teacher models are frozen to guide the training of the student model with the proposed knowledge distillation approach, which follows the same training pipeline as that of the teacher models.

\textbf{Inference.}
In line with common practices, Hanning window penalties are applied during inference to integrate positional prior on tracking \cite{zhang2020ocean}. Specifically, a Hanning window of the same size is multiplied by the classification map, and the location with the highest score is selected as the predicted result.

    

\begin{figure*}[t]
    \centering    \includegraphics[width=0.32\textwidth,height=0.25\textwidth]{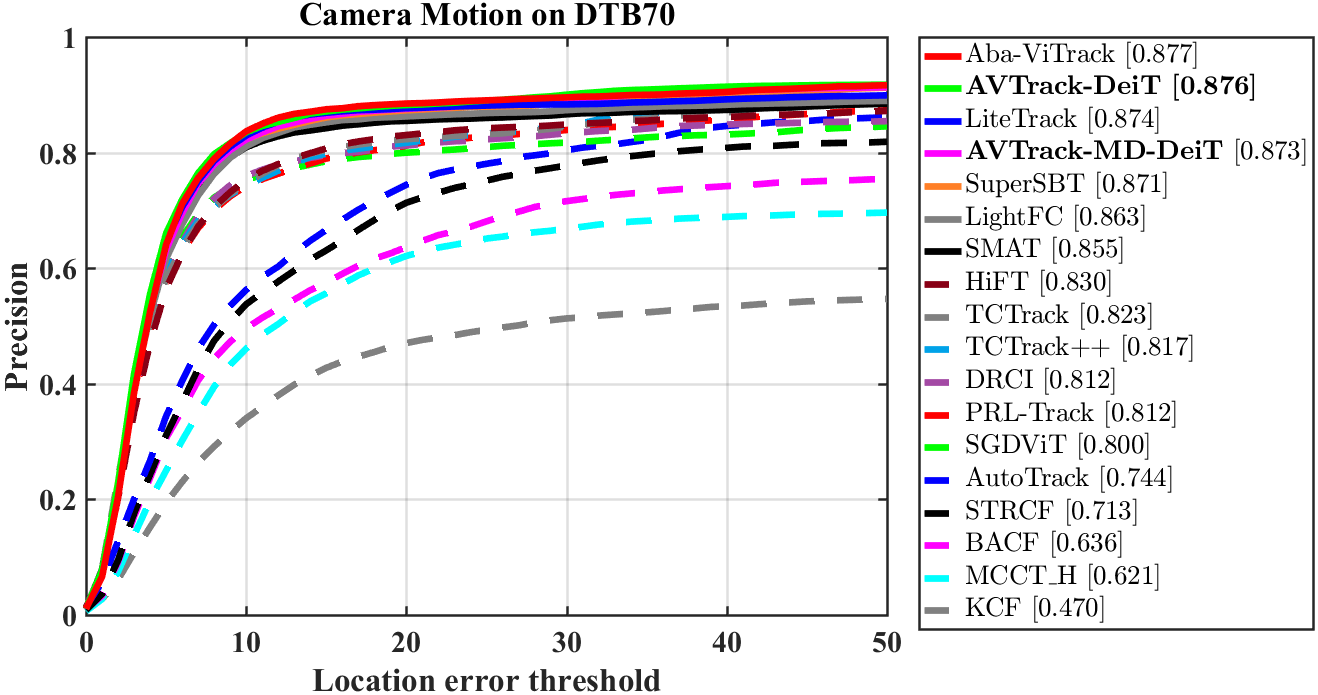}
    \includegraphics[width=0.32\textwidth,height=0.25\textwidth]{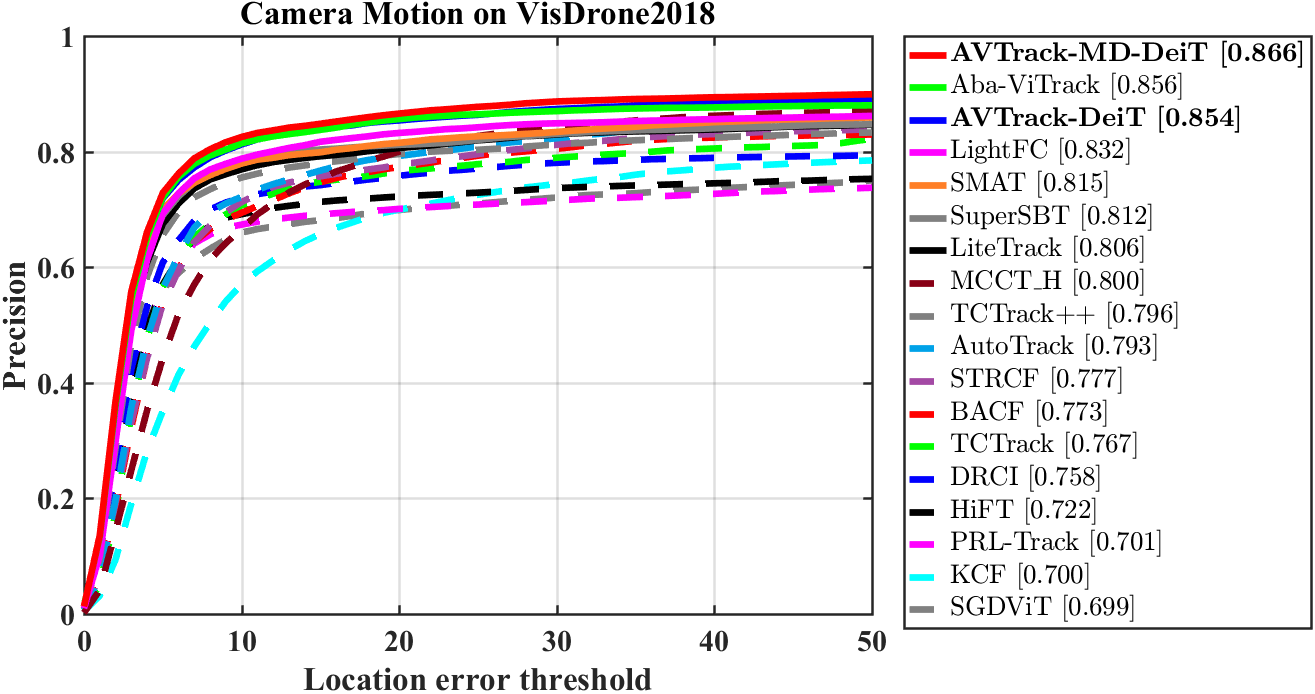}
    \includegraphics[width=0.32\textwidth,height=0.25\textwidth]{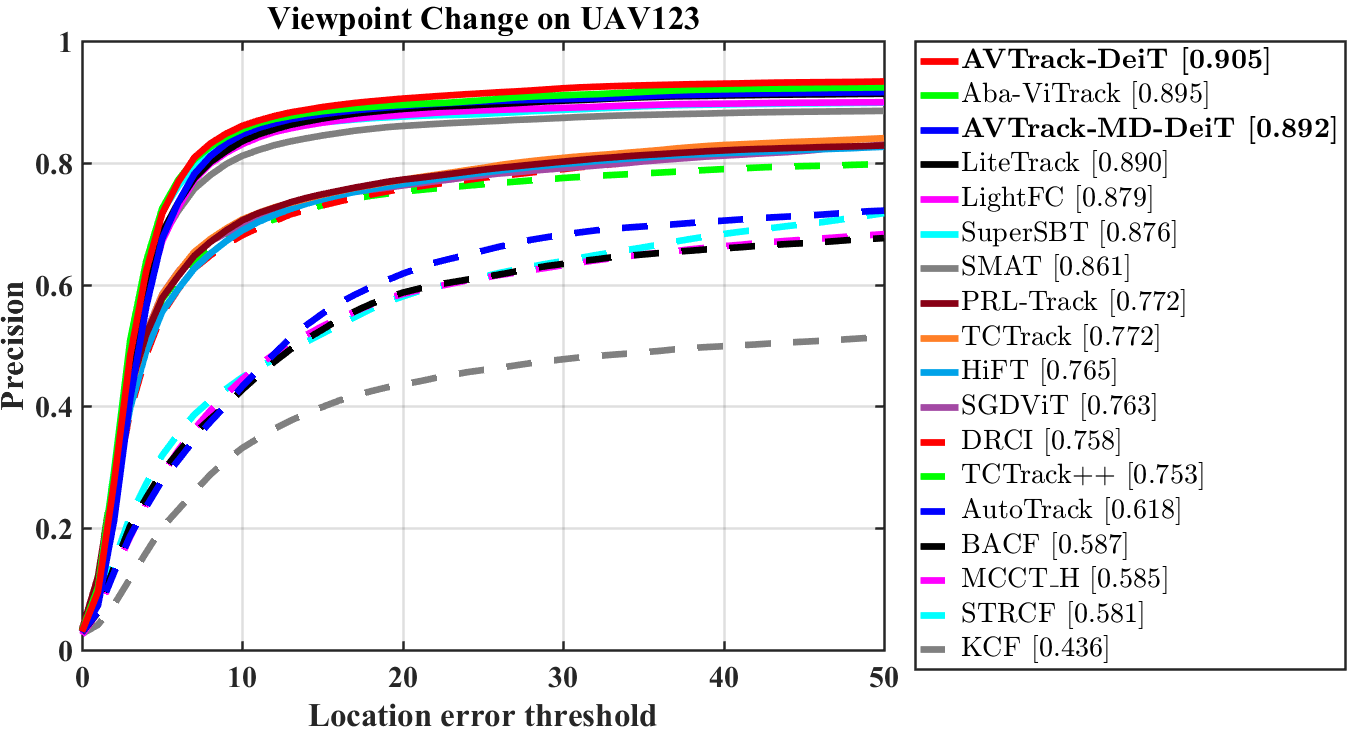}


    \includegraphics[width=0.32\textwidth,height=0.25\textwidth]{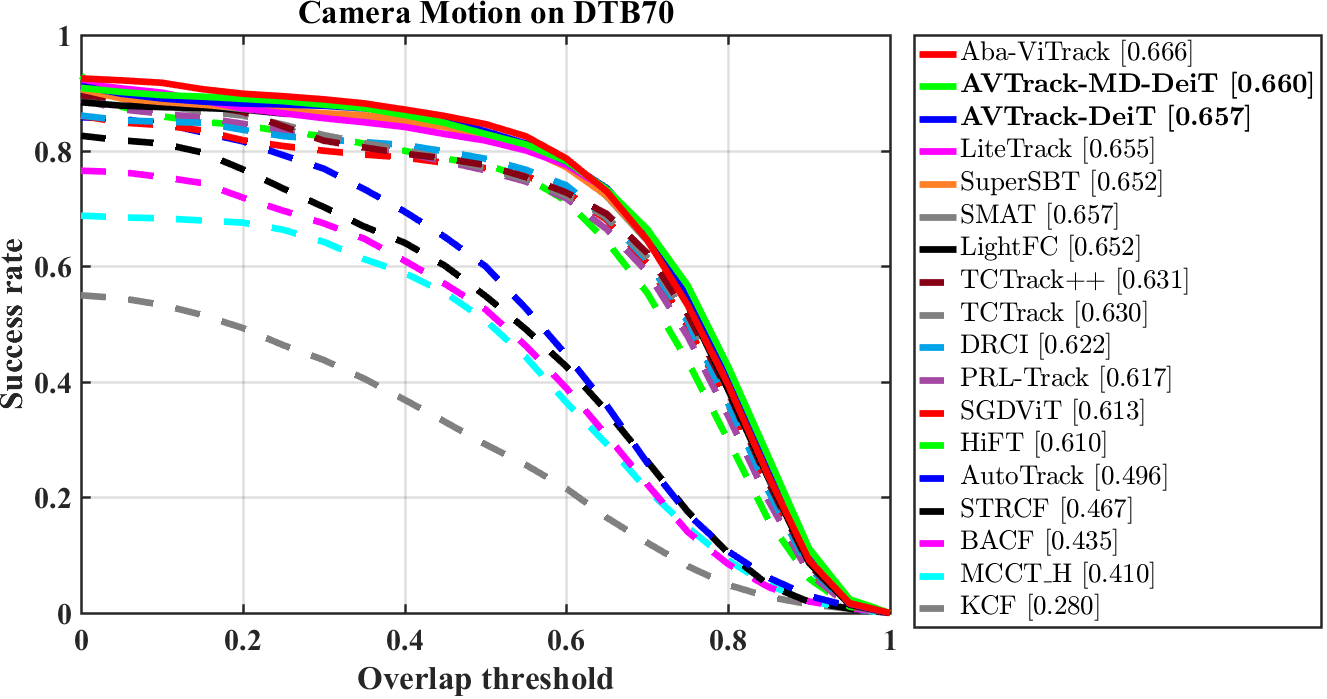}
    \includegraphics[width=0.32\textwidth,height=0.25\textwidth]{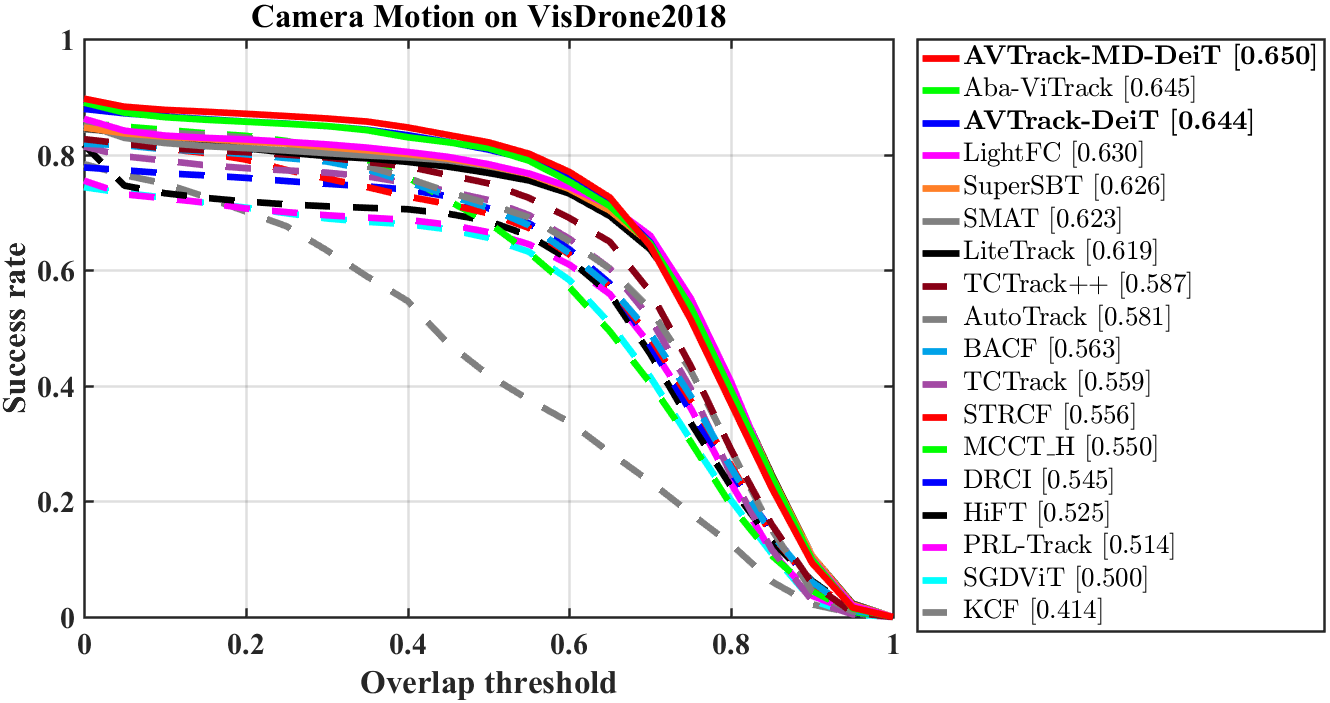}
    \includegraphics[width=0.32\textwidth,height=0.25\textwidth]{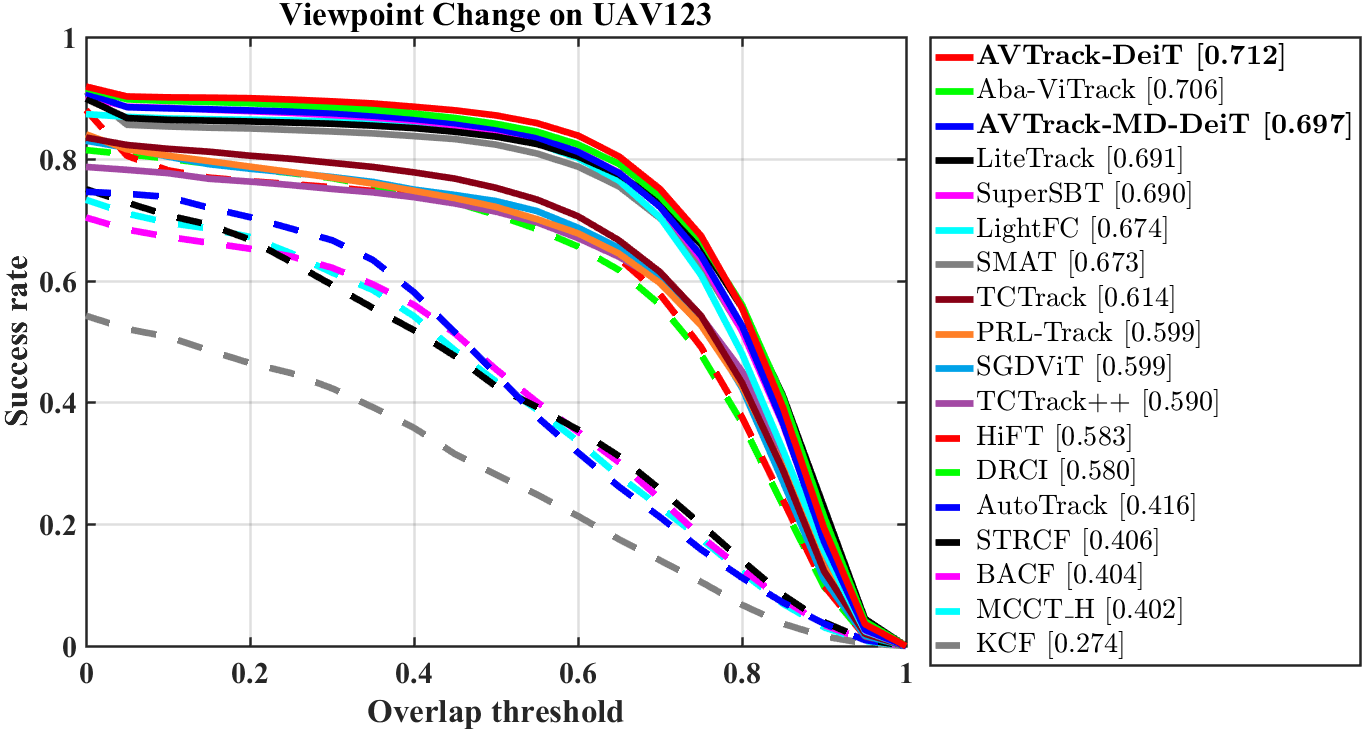}
\caption{The precision plots and success plots of attribute-based comparison are presented for the attribute subsets of DTB70 (the first column), VisDrone2018 (the middle column), and UAV123 (the last column).}
\label{fig_attr_plots}
\end{figure*}

\subsection{Comparison with Lightweight Trackers}
1) \textbf{Overall Performance}: We compare our methods with 22 SOTA lightweight trackers across five renowned UAV tracking datasets.
The evaluation results are presented in Table \ref{table_lightweight_trackers}.
From table, our AVTrack-DeiT~\cite{lilearning2024} outperforms all SOTA trackers except Aba-ViTrack \cite{li2023adaptive} across these five benchmarks in terms of average (Avg.) precision (Prec.) and success rate (Succ.).
Specifically, RACF \cite{li2022learning} exhibits the best performance among DCF-based trackers, with Avg. Prec. and Succ. of 74.6\% and 51.2\%, respectively.
Among CNN-based trackers, DRCI \cite{zeng2023towards} stands out with the highest Avg. Prec. and Succ. of 79.8\% and 59.1\%.
However, even the best methods among DCF- and CNN-based trackers fall short of the worst method among ViT-based trackers. All ViT-based trackers achieve Avg. Prec. and Succ. exceeding 80.0\% and 62.0\%, respectively, clearly surpassing the CNN-based methods and significantly outperforming the DCF-based ones.
Regarding GPU speed, the top three trackers are our proposed methods: AVTrack-MD-EVA, AVTrack-MD-DeiT, and AVTrack-MD-ViT, achieving tracking speeds of 334.4 FPS, 310.6 FPS, and 303.1 FPS, respectively.
In terms of CPU speed, all our methods deliver real-time performance on a single CPU\footnote{It is important to note that the real-time performance discussed in this work is applicable only to platforms that are similar to or more advanced than ours.}.
Obviously, DCF-based trackers are the most efficient UAV trackers, as all methods achieving speeds above 80 FPS belong to this category.
However, these fastest DCF-based trackers are unable to achieve performance comparable to our methods. For instance, the fastest KCF~\cite{henriques2015high} tracker achieves an average success rate of only 31.6\%, approximately half of the performance attained by our methods.
Additionally, DCF-based methods often incur substantial costs to enhance tracking precision.
Specifically, RACF exhibits the best performance among DCF-based trackers; however, it runs at only 35.6 FPS, which is significantly lower than the slowest CPU speed of our trackers.
Although CNN-based trackers can achieve tracking speeds comparable to our ViT-based trackers, the latter significantly outperform the former in overall performance.
Although Aba-ViTrack achieves the highest average precision of 85.3\% and the highest average success rate of 64.7\%., our AVTrack-DeiT secures second place with only slight gaps of 1.2\% and 0.3\%, respectively.
Furthermore, our AVTrack-DeiT demonstrates higher speeds than Aba-ViTrack.
Notably, AVTrack-DeiT runs up to 41\%/18\% faster than Aba-ViTrack in terms of GPU/CPU speed.
When the proposed MI-based multi-teacher knowledge distillation framework (MD) is applied to AVTrack-*, the resulting student model, AVTrack-MD-*, demonstrates a smaller performance drop, with some even achieving performance improvements, all while delivering a significant speed boost.
Specifically, AVTrack-MD-* models show varying trade-offs: AVTrack-MD-ViT achieves 0.5\% and 0.9\% gains in Avg. Prec. and Succ., respectively, with 21.1\% GPU and 8.7\% CPU speed improvements; AVTrack-MD-EVA gains 0.4\% in precision and a minor 0.1\% drop in success rate, while also improving GPU and CPU speeds by 17.9\% and 6.8\%; AVTrack-MD-DeiT has a slight performance drop of 0.5\% in precision and 0.3\% in success rate but still benefits from significant 20.9\% GPU and 8.9\% CPU speed gains.
These results highlight the advantages of our method and justify its SOTA performance for UAV tracking.

\begin{table}[h]
\centering
\scriptsize
\setlength\tabcolsep{3.5pt}
\caption{Comparison of FLOPs, parameters, inference speed (AGX.FPS) on the NVIDIA Jetson AGX Xavier edge device, and performance on WebUAV-3M \cite{zhang2022webuav}.}
\label{table_webuav3M_paramaters}
\begin{tabular}{cccccc}
\toprule[1pt]
Tracker                & Prec.                                & Succ.                                & Params.(M) & FLOPs(G) & AGX.FPS                              \\ \hline
\rowcolor[HTML]{D9D9D9}\textbf{AVTrack-DeiT}    & {\color[HTML]{3531FF} \textBF{70.0}} & {\color[HTML]{FE0000} \textBF{56.4}} & 3.5-7.9    & 0.97-2.4 & {\color[HTML]{3531FF} \textBF{42.3}} \\
\rowcolor[HTML]{eff7ff}\textBF{AVTrack-MD-DeiT} & {\color[HTML]{009901} \textBF{69.4}} & {\color[HTML]{3531FF} \textBF{55.3}} & 5.3        & 1.5      & {\color[HTML]{FE0000} \textBF{46.1}} \\
PRL-Track~\cite{fu2024progressive}       & 62.3                                 & 47.1                                 & 12.1       & 7.4      & 33.8                                 \\
LiteTrack~\cite{wei2024litetrack}       & 69.1                                 & {\color[HTML]{009901} \textBF{54.1}} & 28.3       & 7.3      & 32.1                                 \\
SMAT~\cite{gopal2024separable}            & 68.9                                 & 53.9                                 & 8.6        & 3.2      & 32.7                                 \\
Aba-ViTrack~\cite{li2023adaptive}     & {\color[HTML]{FE0000} \textBF{70.4}} & {\color[HTML]{3531FF} \textBF{55.3}} & 8.0        & 2.4      & {\color[HTML]{009901} \textBF{37.3}} \\
ABDNet~\cite{zuo2023adversarial}          & 63.9                                 & 48.7                                 & 12.3       & 8.3      & 33.2                                 \\
SGDViT~\cite{yao2023sgdvit}          & 61.3                                 & 45.7                                 & 23.3       & 11.3     & 31.7                                 \\
TCTrack++~\cite{cao2023towards}       & 63.9                                 & 48.3                                 & 17.6       & 8.8      & 32.5                                 \\
TCTrack~\cite{cao2022tctrack}         & 61.9                                 & 45.7                                 & 8.5        & 6.9      & 34.4                                 \\
SiamAPN~\cite{fu2021siamese}         & 62.5                                 & 45.1                                 & 14.5       & 7.9      & 35.2                                 \\
HiFT~\cite{cao2021hift}            & 60.9                                 & 45.5                                 & 9.9        & 7.2      & 35.6  \\ \bottomrule[1pt]                              
\end{tabular}
\end{table}

2) \textbf{Performance on WebUAV-3M~\cite{zhang2022webuav}}:
As illustrated in Table \ref{table_webuav3M_paramaters}, we have conducted a comparison of our AVTrack-DeiT and AVTrack-MD-DeiT with ten lightweight SOTA trackers on WebUAV-3M.
Our AVTrack-DeiT surpasses all other lightweight trackers in success rate, achieving a 1.1\% improvement over the second-place Aba-ViTrack while maintaining comparable precision with only a 0.4\% difference. 
Notably, our AVTrack-MD-DeiT experiences less than a 1.0\% drop in both precision and success rate, while achieving a 9.2\% improvement in AGX speed.
These results further underscore the effectiveness of our approach.

3) \textbf{Efficiency Comparison}:
To further demonstrate that our tracker achieves a better trade-off between accuracy and efficiency, we compare our methods against ten SOTA lightweight trackers, in terms of floating point operations (FLOPs), the number of parameters (Params.), and the speed (AGX.FPS) on the Nvidia Jetson AGX Xavier edge device.
The results are shown in Table \ref{table_webuav3M_paramaters}.
Notably, since our AVTrack-DeiT tracker features adaptive architectures, the FLOPs and parameters range from minimum to maximum values.
As observed, both the minimum FLOPS and Params. of our AVTrack-DeiT are notably lower than those of all the SOTA trackers, and even its maximum values are lower than those of most SOTA trackers.
Furthermore, our AVTrack-MD-DeiT achieves the lowest FLOPs (1.5G) and parameters (5.3M), except for AVTrack-DeiT, while also delivering the fastest speed at 46.1 AGX.FPS.
This comparison in terms of computational complexity also underscores the efficiency of our methods.

\begin{table}[!h]
\centering
\scriptsize
\setlength\tabcolsep{5.5pt}
\caption{Compwith state-of-the-art methods on two long-term UAV tracking benchmarks: UAV20L~\cite{mueller2016benchmark} and UAVTrack112L~\cite{fu2021onboard}.}
\label{table:long-term}
\begin{tabular}{ccccc}
\bottomrule[1pt]
                          & \multicolumn{2}{c}{UAV20L}                                                  & \multicolumn{2}{c}{UAVTrack112L}                                            \\
\multirow{-2}{*}{Tracker} & Prec.                                & Succ.                                & Prec.                                & Succ.                                \\ \hline
\rowcolor[HTML]{D9D9D9}\textbf{AVTrack-DeiT}     & {\color[HTML]{FE0000} \textbf{85.6}} & {\color[HTML]{FE0000} \textbf{68.2}} & {\color[HTML]{3531FF} \textbf{80.2}} & {\color[HTML]{3531FF} \textbf{63.9}} \\
\rowcolor[HTML]{eff7ff}\textbf{AVTrack-MD-DeiT}  & 83.8                                 & 66.6                                 & {\color[HTML]{009901} \textbf{78.8}} & {\color[HTML]{009901} \textbf{62.7}} \\
PRL-Track~\cite{fu2024progressive}                 & 71.8                                 & 57.1                                 & 78.4                                 & 59.7                                 \\
LiteTrack~\cite{wei2024litetrack}                 & {\color[HTML]{009901} \textbf{84.8}} & {\color[HTML]{009901} \textbf{66.7}} & 78.8                                 & 62.1                                 \\
SMAT~\cite{gopal2024separable}                      & 82.3                                 & 65.9                                 & 76.4                                 & 60.8                                 \\
Aba-ViTrack~\cite{li2023adaptive}               & {\color[HTML]{3531FF} \textbf{85.2}} & {\color[HTML]{3531FF} \textbf{67.9}} & {\color[HTML]{FE0000} \textbf{81.1}} & {\color[HTML]{FE0000} \textbf{64.2}} \\
ABDNet~\cite{zuo2023adversarial}                    & 74.4                                 & 59.5                                 & 77.1                                  & 60.2                                  \\
SGDViT~\cite{yao2023sgdvit}                    & 69.7                                 & 55.8                                 & 73.8                                 & 55.1                                 \\
TCTrack++~\cite{cao2023towards}                 & 74.0                                 & 58.8                                 & 78.7                                 & 60.0                                 \\
TCTrack~\cite{cao2022tctrack}                   & 75.1                                 & 59.6                                 & 72.0                                 & 53.7                                 \\
SiamAPN\cite{fu2021siamese}                   & 70.9                                 & 55.6                                 & 78.0                                 & 57.8                                 \\
HiFT~\cite{cao2021hift}                      & 78.0                                 & 56.8                                 & 70.8                                 & 53.1  \\ \bottomrule[1pt]                              
\end{tabular}
\end{table}

4) \textbf{Performance in Long-Term Tracking Scenarios}: To better demonstrate the effectiveness of the proposed methods in long-term tracking, we have evaluated the performance of the proposed tracker on two long-term UAV tracking benchmarks: UAV20L~\cite{mueller2016benchmark} and UAVTrack112L~\cite{fu2021onboard}, as shown in Table \ref{table:long-term}. 
As observed, our AVTrack-DeiT achieves the best performance on UAV20L and the second-best on UAVTrack112L. Additionally, AVTrack-MD-DeiT performs comparably to AVTrack-DeiT and ranks third on UAVTrack112L. Although Aba-ViTrack achieves performance comparable to our trackers, as shown in Table \ref{table_lightweight_trackers}, its tracking speed is significantly slower than that of our methods.
The results demonstrate that our tracker achieves competitive performance compared to the state-of-the-art methods, confirming its robustness in long-term tracking scenarios.

5) \textbf{Attribute-Based Evaluation}:
We also performed attribute-based evaluations to assess the robustness of our trackers (i.e., AVTrack-DeiT and AVTrack-MD-DeiT) in challenging scenarios involving viewpoint changes.
This includes comparing our trackers with 16 SOTA lightweight trackers under challenges such as `Camera Motion' on the DTB70 and VisDrone2018 and `Viewpoint Change' on the UAV123, as shown in Fig. \ref{fig_attr_plots}.
Notably, the DTB70 and VisDrone2018 datasets do not have a dedicated subset for scenarios involving viewpoint changes. As an alternative, we used the `Camera Motion' subset to evaluate performance in such scenarios, as it is a key factor contributing to viewpoint changes in visual tracking.
For instance, when a camera or the object moves rapidly or changes direction suddenly, it can result in varying views of the target in captured images.
As observed, our tracker ranks first in `Camera Motion' on VisDrone2018, and `Viewpoint Change' on UAV123, exceeding the second-best tracker by 1.0\%/0.5\% and 1.0\%/0.6\% in precision/success rate, respectively, with only slight gaps of 0.1\% and 0.6\% to the top tracker in the `Camera Motion' on DTB70.
Therefore, this attribute-based evaluation validates, both directly and indirectly, the superiority and effectiveness of the proposed method in addressing the challenges associated with viewpoint change.

\begin{figure}[h]
\centering
\includegraphics[width=0.475\textwidth]{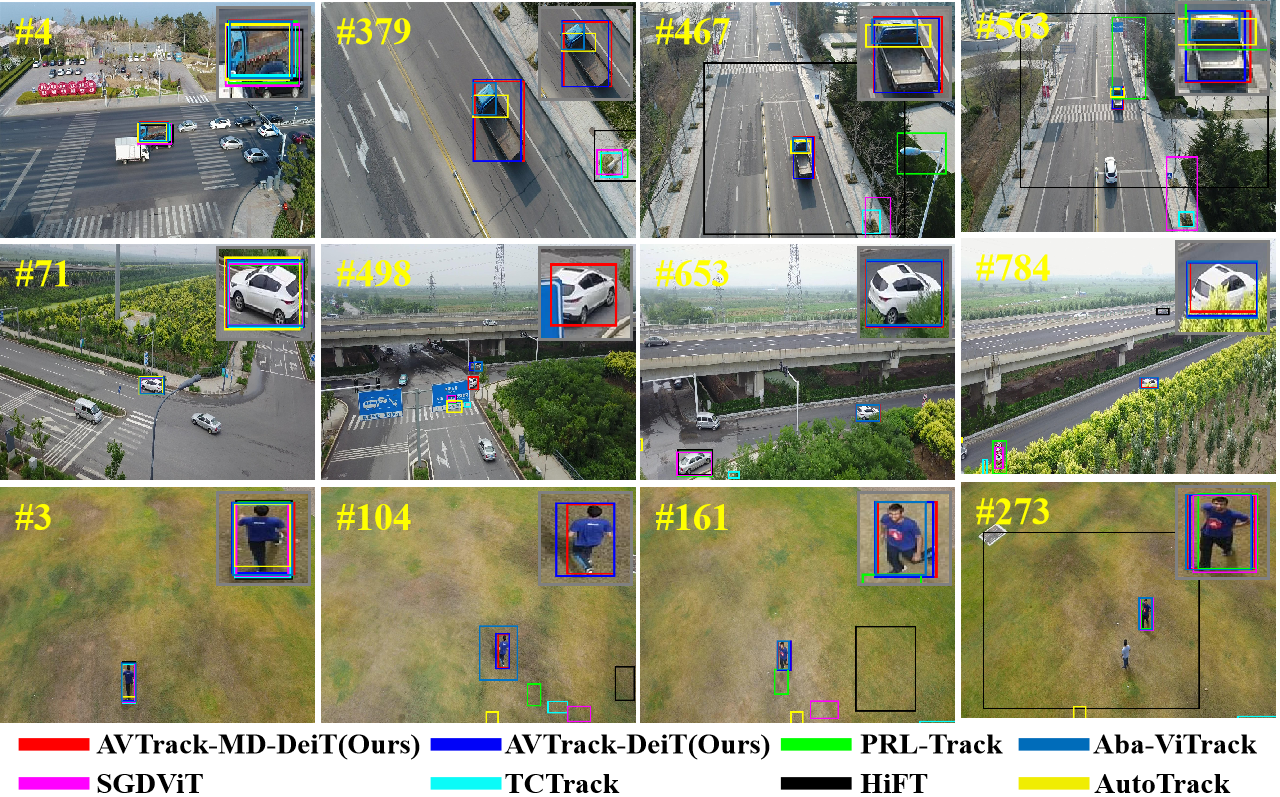}
\caption{The qualitative evaluations were performed on three video sequences from the following datasets: UAVDT \cite{du2018the}, VisDrone2018 \cite{wen2018visdrone}, and UAV123@10fps \cite{mueller2016benchmark}. The evaluated sequences include S1607, uav0000180\_00050\_s, and person10.}
\label{fig_bbox}
\end{figure}

4) \textbf{Qualitative evaluation}:
To intuitively showcase the tracking results in UAV tracking scenes, qualitative tracking results of our trackers (i.e., AVTrack-MD-DeiT and AVTrack-DeiT) alongside six SOTA UAV trackers are visualized in Fig. \ref{fig_bbox}.
Three video sequences, selected from different benchmarks and scenarios, are presented for demonstration: S1607 from UAVDT, uav0000180\_00050\_s from VisDrone2018, and person10 from UAV123@10fps.
The upper right corner features a selectively zoomed and cropped view of the tracked objects within the corresponding frames for better visualization.
As observed, compared to the other SOTA UAV trackers, our tracker tracks the target objects more accurately in all challenging scenes, including occlusion (i.e., S1607 and uav0000180\_00050\_s), scale variations (i.e., in all sequences), and viewpoint change (i.e., in all sequences).
In these cases, our method significantly outperforms others and offers a more visually appealing result, highlighting the effectiveness of the proposed approaches for UAV tracking.

\begin{table*}[t]
\scriptsize
\centering
\setlength\tabcolsep{3.5pt}
\caption{Precision (Prec.), success (Succ.), and average GPU speed (Avg.FPS) comparison between our proposed trackers(i.e., AVTrack-DeiT and AVTrack-MD-DeiT) and DL-based tracker on five UAV tracking benchmarks.}
\label{table_deep_tracker}
\begin{tabular}{ccccccccccccccc}
\hline
                          &                              & \multicolumn{2}{c}{DTB70}                                                   & \multicolumn{2}{c}{UAVDT}                                                   & \multicolumn{2}{c}{VisDrone2018}                                            & \multicolumn{2}{c}{UAV123}                                                  & \multicolumn{2}{c}{UAV123@10fps}                                            & \multicolumn{2}{c}{Avg.}                                                    &                                       \\
\multirow{-2}{*}{Tracker} & \multirow{-2}{*}{Source}     & Prec.                                & Succ.                                & Prec.                                & Succ.                                & Prec.                                & Succ.                                & Prec.                                & Succ.                                & Prec.                                & Succ.                                & Prec.                                & Succ.                                & \multirow{-2}{*}{Avg.FPS}             \\ \hline
DiMP\cite{bhat2019learning}                      & ICCV 19                      & 79.2                                 & 61.3                                 & 78.3                                 & 57.4                                 & 83.5                                 & 63.0                                 & 83.1                                 & 65.2                                 & 85.1                                 & 64.7                                 & 81.8          & 62.3                                 & 51.9           \\
PrDiMP\cite{Danelljan2020ProbabilisticRF}                      & CVPR 20                      & 84.0                                 & 64.3                                 & 75.8                                 & 55.9                                 & 79.8                                 & 60.2                                 & 87.2                                 & 66.5                                 & 83.9                                 & 64.7                                 & 82.1                                 & 62.3                                 & 53.6                                  \\
TrSiam\cite{wang2021transformer}                    & CVPR 21                      & 82.7                                 & 63.9                                 & {\color[HTML]{FE0000} \textbf{88.9}} & {\color[HTML]{3531FF} \textbf{65.0}} & 84.0                                 & 63.5                                 & 83.9                                 & 66.3                                 & 85.3                                 & 64.9                                 & 84.9                                 & 64.7                                 & 38.3                                  \\
TranT\cite{chen2021transformer}                    & CVPR 21                  & 83.6                                 & 65.8                                 & 82.6                                 & 63.2                                 & 85.9                                 & 65.2                                 & 85.0                                 & 67.1                                 & 84.8
& 66.5
& 84.4
& 65.6                                 & 53.2                                  \\
AutoMatch\cite{Zhang2021LearnTM}                 & ICCV 21                      & 82.5                                 & 63.4                                 & 82.1                                 & 62.9                                 & 78.1                                 & 59.6                                 & 83.8                                 & 64.4                                 & 78.1                                 & 59.4                                 & 80.9                                 & 61.9                                 & 63.9                                  \\
SparseTT\cite{2022SparseTT}                  & IJCAI 22                     & 82.3                                 & 65.8                                 & 82.8                                 & {\color[HTML]{FE0000} \textbf{65.4}} & 81.4                                 & 62.1                                 & 85.4                                 & 68.8                                 & 82.2                                 & 64.9                                 & 82.8                                 & 65.4                                 & 32.4                                  \\
CSWinTT\cite{song2022transformer}                   & CVPR 22                      & 80.3                                 & 62.3                                 & 67.3                                 & 54.0                                 & 75.2                                 & 58.0                                 & 87.6                                 & {\color[HTML]{FE0000} \textbf{70.5}} & 87.1                                 & 68.1                                 & 79.5                                 & 62.6                                 & 12.3                                  \\
SimTrack\cite{chen2022backbone}                  & ECCV 22                      & 83.2                                 & 64.6                                 & 76.5                                 & 57.2                                 & 80.0                                 & 60.9                                 & 88.2                                 & 69.2                                 & 87.5                                 & 69.0                                 & 83.1                                 & 64.2                                 & {\color[HTML]{009901} \textbf{72.8}}  \\
OSTrack\cite{ye2022joint}                   & ECCV 22                      & 82.7                                 & 65.0                                 & {\color[HTML]{3531FF} \textbf{85.1}} & {\color[HTML]{009901} \textbf{63.4}} & 84.2                                 & 64.8                                 & 84.7                                 & 67.4                                 & 83.1                                 & 66.1                                 & 83.9                                 & 65.3                                 & 68.4                                  \\
ZoomTrack\cite{kou2024zoomtrack}                 & NIPS 23                      & 82.0                                 & 63.2                                 & 77.1                                 & 57.9                                 & 81.4                                 & 63.6                                 & {\color[HTML]{009901} \textbf{88.4}} & {\color[HTML]{009901} \textbf{69.6}} & {\color[HTML]{3531FF} \textbf{88.8}} & {\color[HTML]{009901} \textbf{70.0}} & 83.5                                 & 64.8                                 & 62.7                                  \\
SeqTrack\cite{Chen2023SeqTrackST}                  & CVPR 23                      & 85.6                                 & 65.5                                 & 78.7                                 & 58.8                                 & 83.3                                 & 64.1                                 & 86.8                                 & 68.6                                 & 85.7                                 & 68.1                                 & 84.0                                 & 65.0                                 & 32.3                                  \\
MAT\cite{zhao2023representation}                       & CVPR 23                      & 83.2                                 & 64.5                                 & 72.9                                 & 54.8                                 & 81.6                                 & 62.2                                 & 86.7                                 & 68.3                                 & 86.9                                 & 68.5                                 & 82.3                                 & 63.6                                 & 71.2           \\
ROMTrack\cite{cai2023robust}                  & ICCV 23                      & {\color[HTML]{3531FF} \textbf{87.2}} & {\color[HTML]{3531FF} \textbf{67.4}} & 81.9                                 & 61.6                                 & {\color[HTML]{3531FF} \textbf{86.4}} & {\color[HTML]{3531FF} \textbf{66.7}} & 87.4                                 & 69.2                                 & 85.0                                 & 67.8                                 & {\color[HTML]{009901} \textbf{85.5}} & {\color[HTML]{009901} \textbf{66.5}} & 52.3                                  \\
DCPT\cite{zhu2024dcpt}                      & ICRA 24                      & 84.0                                 & 64.8                                 & 76.8                                 & 56.9                                 & 83.1                                 & 64.2                                 & 85.7                                 & 68.1                                 & 86.9                                 & 69.1                                 & 83.3                                 & 64.6                                 & 39.2                                  \\
HIPTrack\cite{cai2024hiptrack}                  & CVPR 24                      & {\color[HTML]{FE0000} \textbf{88.4}} & {\color[HTML]{FE0000} \textbf{68.6}} & 79.6                                 & 60.9                                 & {\color[HTML]{FE0000} \textbf{86.7}} & {\color[HTML]{FE0000} \textbf{67.1}} & {\color[HTML]{FE0000} \textbf{89.2}} & {\color[HTML]{FE0000} \textbf{70.5}} & {\color[HTML]{FE0000} \textbf{89.3}} & {\color[HTML]{FE0000} \textbf{70.6}} & {\color[HTML]{FE0000} \textbf{86.6}} & {\color[HTML]{FE0000} \textbf{67.5}} & 32.1                                  \\
EVPTrack\cite{shi2024evptrack}                  & AAAI 24                      & {\color[HTML]{009901} \textbf{85.8}} & {\color[HTML]{009901} \textbf{66.5}} & 80.6                                 & 61.2                                 & 84.5                                 & {\color[HTML]{009901} \textbf{65.8}} & {\color[HTML]{3531FF} \textbf{88.9}} & {\color[HTML]{3531FF} \textbf{70.2}} & {\color[HTML]{009901} \textbf{88.7}} & {\color[HTML]{3531FF} \textbf{70.4}} & {\color[HTML]{3531FF} \textbf{85.7}} & {\color[HTML]{3531FF} \textbf{66.8}} & 26.1                                  \\
\rowcolor[HTML]{D9D9D9} 
\textbf{AVTrack-DeiT}              & \textbf{Ours}                         & 84.3                                 & 65.0                                 & 82.1          & 58.7          & {\color[HTML]{009901} \textbf{86.0}} &  65.3          & 84.8          & 66.8          &  83.2          & 65.8          &  84.1          & 64.4          & {\color[HTML]{3531FF} \textbf{256.8}} \\
\rowcolor[HTML]{eff7ff} 
\textbf{AVTrack-MD-DeiT}           & \cellcolor[HTML]{eff7ff} \textbf{Ours} & 84.0                                 & 65.2          & {\color[HTML]{009901} \textbf{83.1}} & 60.3          & 84.9                                 & 64.2                                 & 82.6                                 & 65.2                                 & 83.3                                 & 65.5                                 & 83.6                                 & 64.1          & {\color[HTML]{FE0000} \textbf{310.6}} \\ \hline
\end{tabular}
\end{table*}

\subsection{Comparison with Deep Trackers}

To further validate the superiority of our trackers, we compare AVTrack-DeiT and AVTrack-MD-DeiT with 16 deep SOTA trackers across five UAV tracking datasets.
The evaluation results are shown in Table \ref{table_deep_tracker}, which shows precision (Prec.), success rate (Succ.), their average (Avg.), and average GPU speed (Avg.FPS).
As shown, our AVTrack-MD-DeiT and AVTrack-DeiT excel by achieving the fastest and second-fastest speeds while maintaining high performance comparable to the SOTA deep trackers. 
On average, although AVTrack-MD-DeiT and AVTrack-DeiT exhibit slight gaps in average precision and success rate compared to HIPTrack~\cite{cai2024hiptrack}, EVPTrack~\cite{shi2024evptrack}, and ROMTrack~\cite{cai2023robust}, they are noticeably slower than our methods in terms of GPU speed.
Specifically, our AVTrack-MD-DeiT is 9.7, 11.9, and 5.9 times faster than HIPTrack, EVPTrack, and ROMTrack, respectively.
These results indicate that our method delivers both high precision and speed, validating its suitability for real-time UAV tracking that prioritizes efficiency as well as accuracy.

\begin{table*}[t]
\scriptsize
\setlength\tabcolsep{3.0pt} 
\centering
\caption{Comparison of original teacher models and distillation models with varying teacher combinations: model complexity, performance, and average GPU speed. $\psi_1$, $\psi_2$, and $\psi_3$ represent the AVTrack-DeiT, AVTrack-ViT, and AVTrack-EVA, respectively.}
\label{table_abla_varis_teachers}
\begin{tabular}{c|ccc|cc|cc|cc|cc|cc|cc|cc|c}
\toprule[1pt]
\multirow{2}{*}{Model}        & \multirow{2}{*}{$\psi_1$} & \multirow{2}{*}{$\psi_2$} & \multirow{2}{*}{$\psi_3$} & \multirow{2}{*}{Params} & \multirow{2}{*}{FLOPS} & \multicolumn{2}{c|}{DTB70}     & \multicolumn{2}{c|}{UAVDT}     & \multicolumn{2}{c|}{VisDrone}  & \multicolumn{2}{c|}{UAV123}    & \multicolumn{2}{c|}{UAV123@10fps} & \multicolumn{2}{c|}{Avg.}       & \multirow{2}{*}{Avg.FPS} \\
                              &                     &                     &                     &                         &                        & Prec.         & Succ.         & Prec.         & Succ.         & Prec.         & Succ.         & Prec.         & Succ.         & Prec.           & Succ.          & Prec.         & Succ.         &                          \\ \hline
AVTrack-DeiT                  & -                   & -                   & -                   & 3.5-7.9                 & 0.97-2.4               & 84.3          & 65.0          & 82.1          & 58.7          & 86.0          & 65.3          & 84.8          & 66.8          & 83.2            & 65.8           & 84.1          & 64.4          & 256.8                    \\ 
AVTrack-ViT                   & -                   & -                   & -                   & 3.5-8.0                 & 0.97-2.4               & 81.3          & 63.3          & 79.9          & 57.7          & 86.4          & 65.9          & 84.0          & 66.2          & 83.2            & 65.7           & 82.9          & 63.1          & 250.2                    \\
AVTrack-EVA                   & -                   & -                   & -                   & 2.4-5.8                 & 0.67-1.7               & 82.6          & 64.0          & 78.8          & 57.2          & 84.4          & 63.5          & 83.0          & 64.7          & 81.2            & 63.5           & 82.0          & 62.6          & 283.7                    \\ \hline
\multirow{3}{*}{AVTrack-MD-DeiT} & $\checkmark$                   &-                     & -                    & \multirow{3}{*}{5.3}    & \multirow{3}{*}{1.5}   & 81.5	& 62.9	& 80.5	& 57.6	& 82.5	& 62.4	& 81.8	& 64.1	& 81.1	& 63.7	& 81.7$_{\downarrow 2.6\%}$	& 62.1$_{\downarrow 2.3\%}$          &                         \\
                              & $\checkmark$                   & $\checkmark$                   &-                     &                         &                        & 83.4	& 64.8	& 80.9	& 58.1	& 83.4	& 62.9	& 82.5	& 64.9	& 82.4	& 64.8	& 82.5$_{\downarrow 1.6\%}$	& 63.1$_{\downarrow 1.3\%}$         & 310.6$_{\uparrow 20.9\%}$                        \\ 
                              & $\checkmark$                   & $\checkmark$                   & $\checkmark$                   &                         &                        & \textbf{84.0} & \textbf{65.2} & \textbf{83.1} & \textbf{60.3} & \textbf{84.9} & \textbf{64.2} & \textbf{82.6}          & \textbf{65.2}          & \textbf{83.3}            & \textbf{65.5}           & \textbf{83.6}$_{\downarrow 0.5\%}$ & \textbf{64.1}$_{\downarrow 0.3\%}$ &  \\ \hline
\multirow{3}{*}{AVTrack-MD-ViT}  & $\checkmark$                   &-                     &-                     & \multirow{3}{*}{5.3}    & \multirow{3}{*}{1.5}   & 79.1	& 61.2	& 77.9	& 55.8	& 82.1	& 62.2 & 81.8	& 63.8	& 81.0	& 62.5		& 80.3$_{\downarrow 2.6\%}$	& 61.1$_{\downarrow 2.0\%}$          &                         \\
                              & $\checkmark$                   & $\checkmark$                   & -                     &                         &                        & 81.5	& 62.9	& 80.4	& 58.4	& 81.8	& 61.7	& 82.1	& 64.2  & \textbf{83.9}	& \textbf{65.9}		& 81.9$_{\downarrow 1.0\%}$ & 62.6$_{\downarrow 0.5\%}$          & 303.1$_{\uparrow 21.1\%}$                        \\
                              & $\checkmark$                   & $\checkmark$                   & $\checkmark$                   &                         &                        & \textbf{84.9} & \textbf{65.7} & \textbf{81.4} & \textbf{59.5} & \textbf{84.8} & \textbf{63.7} & \textbf{82.3}          & \textbf{65.1}          & 83.5   & \textbf{65.9}  & \textbf{83.4}$_{\uparrow 0.6\%}$ & \textbf{64.0}$_{\uparrow 0.9\%}$ &                     \\ \hline
\multirow{3}{*}{AVTrack-MD-EVA}  & $\checkmark$                   & -                     &  -                   & \multirow{3}{*}{3.7}    & \multirow{3}{*}{1.1}   & 80.1	& 61.5	& 77.3	& 55.8	& 80.5	& 61.2 & 80.6	& 62.1	& 80.1	& 61.9		& 79.7$_{\downarrow 2.3\%}$	& 60.5$_{\downarrow 2.1\%}$          &                         \\
      & $\checkmark$                   & $\checkmark$                   & -                     &                         &                        & 81.2	& 62.1	& 79.1	& 57.1	& 81.8	& 62.1	& \textbf{82.1}	& \textbf{63.1}	& 80.4	& 62.1	& 80.9$_{\downarrow 1.1\%}$	& 61.3$_{\downarrow 1.3\%}$          & 334.4$_{\uparrow 17.9\%}$                        \\
                              & $\checkmark$                   & $\checkmark$                   & $\checkmark$                   &                         &                        & \textbf{83.2} & \textbf{63.9} & \textbf{80.8} & \textbf{58.0} & \textbf{84.0} & \textbf{63.5} & 81.5          & 62.3          & \textbf{82.7}            & \textbf{64.7}           & \textbf{82.4}$_{\uparrow 0.4\%}$ & \textbf{62.5}$_{\downarrow 0.1\%}$ &                     \\ \bottomrule[1pt]                   
\end{tabular}
\end{table*}

\subsection{Ablation Study}


\textbf{Impact of the Number of Teachers:}
More teachers may provide more diverse knowledge. 
Table \ref{table_abla_varis_teachers} summarizes the model complexity, performance, and average GPU speed of the original teacher models and the distillation models using different teacher combinations.
As shown, single-teacher distillation can greatly reduce model complexity during inference, resulting in a significant speedup of over 17\%. However, it leads to a considerable decline in performance compared to the corresponding teacher model, with decreases of over 2.0\% in average precision and success rate. In contrast, multi-teacher distillation methods, including double-teacher and triple-teacher distillation, not only reduce model complexity and provide a significant speedup but also result in a smaller performance decrease, with all drops remaining below 1.6\% in average precision and success rate, and some even surpassing the performance of the teacher models.
In the triple-teacher distillation, the student models (i.e., AVTrack-MD-ViT and AVTrack-MD-EVA) exhibit superior performance when compared to their corresponding teacher models with the same backbone.
Specifically, AVTrack-MD-ViT improves average precision by 0.5\% and success rate by 0.9\%, while AVTrack-MD-EVA shows a 0.4\% improvement in average precision with only a 0.1\% drop in success rate.
Although AVTrack-MD-DeiT does not outperform AVTrack-DeiT, it experiences only a slight performance drop, with both average precision and success rate decreasing by less than 0.5\%.
Notably, in practice, the model complexity of AVTrack is higher than that of AVTrack-MD, as the activated blocks in AVTrack typically exceed half of the total blocks in the backbone. As a result, AVTrack is slower than AVTrack-MD when both use the same ViT backbone.
These experimental results provide evidence for the effectiveness of our proposed multi-teacher knowledge distillation based on MI Maximization, with the performance gain attributed to the complementarity of multiple teacher models.

\begin{table*}[t]
\scriptsize
\setlength\tabcolsep{2.5pt} 
\centering
\caption{Performance of varying numbers of ViT blocks in student models across five UAV tracking datasets.}
\label{table_aba_vit_blocks}
\begin{tabular}{c|cc|cc|cc|cc|cc|cc|cc|c}
\toprule[1pt]
& \multicolumn{2}{c|}{DTB70}                                                   & \multicolumn{2}{c|}{UAVDT}                                                   & \multicolumn{2}{c|}{VisDrone}                                                & \multicolumn{2}{c|}{UAV123}                                                  & \multicolumn{2}{c|}{UAV123@10fps}                                            & \multicolumn{2}{c|}{Avg.}                                                    &                                                                        &                                                                       &                                       \\
\multirow{-2}{*}{\begin{tabular}[c]{@{}c@{}}blocks\end{tabular}} & Prec.                                & Succ.                                & Prec.                                & Succ.                                & Prec.                                & Succ.                                & Prec.                                & Succ.                                & Prec.                                & Succ.                                & Prec.                                & Succ.                                & \multirow{-2}{*}{\begin{tabular}[c]{@{}c@{}}Param.\\ (M)\end{tabular}} & \multirow{-2}{*}{\begin{tabular}[c]{@{}c@{}}FLOPs\\ (G)\end{tabular}} & \multirow{-2}{*}{Avg.FPS}             \\ \hline
4                                                                              & 81.7	& 63.2	& 80.2	& 57.2	& 80.8	& 61.0	& 81.6	& 64.3	& 81.1	& 63.8	& 81.1	& 61.9                                 & {\color[HTML]{FE0000} \textBF{4.4}}                                    & {\color[HTML]{FE0000} \textBF{1.3}}                                   & {\color[HTML]{FE0000} \textBF{357.1}} \\
5                                                                              & 82.1                                 & 63.8                                 & 81.9                                 & 58.5                                 & 82.5                                 & 62.1                                 & 82.5                                 & 64.9                                 & {\color[HTML]{009901} \textBF{83.9}} & {\color[HTML]{009901} \textBF{66.0}} & 82.6                                 & 63.1                                 & {\color[HTML]{3531FF} \textBF{4.9}}                                    & {\color[HTML]{3531FF} \textBF{1.4}}                                   & {\color[HTML]{3531FF} \textBF{334.9}} \\
6                                                                              & {\color[HTML]{FE0000} \textBF{84.0}} & {\color[HTML]{FE0000} \textBF{65.2}} & {\color[HTML]{FE0000} \textBF{83.1}} & {\color[HTML]{FE0000} \textBF{60.3}} & {\color[HTML]{3531FF} \textBF{84.9}} & {\color[HTML]{3531FF} \textBF{64.2}} & {\color[HTML]{009901} \textBF{82.6}} & {\color[HTML]{009901} \textBF{65.2}} & 83.3                                 & 65.5                                 & {\color[HTML]{009901} \textBF{83.6}} & {\color[HTML]{009901} \textBF{64.1}} & {\color[HTML]{009901} \textBF{5.3}}                                    & {\color[HTML]{009901} \textBF{1.5}}                                   & {\color[HTML]{009901} \textBF{310.6}} \\
7                                                                              & {\color[HTML]{3531FF} \textBF{83.7}} & {\color[HTML]{3531FF} \textBF{64.6}} & {\color[HTML]{009901} \textBF{82.4}} & {\color[HTML]{009901} \textBF{60.5}} & {\color[HTML]{009901} \textBF{84.8}} & {\color[HTML]{009901} \textBF{63.8}} & {\color[HTML]{3531FF} \textBF{83.7}} & {\color[HTML]{3531FF} \textBF{66.3}} & {\color[HTML]{3531FF} \textBF{84.1}} & {\color[HTML]{3531FF} \textBF{66.3}} & {\color[HTML]{3531FF} \textBF{83.7}} & {\color[HTML]{3531FF} \textBF{64.3}} & 5.8                                                                    & 1.7                                                                   & 291.4                                 \\
8                                                                              & {\color[HTML]{009901} \textBF{82.9}}                                 & {\color[HTML]{009901} \textBF{64.0}}                                 & {\color[HTML]{3531FF} \textBF{82.8}} & {\color[HTML]{3531FF} \textBF{60.8}} & {\color[HTML]{FE0000} \textBF{85.3}} & {\color[HTML]{FE0000} \textBF{64.4}} & {\color[HTML]{FE0000} \textBF{84.1}} & {\color[HTML]{FE0000} \textBF{66.6}} & {\color[HTML]{FE0000} \textBF{84.7}} & {\color[HTML]{FE0000} \textBF{66.7}} & {\color[HTML]{FE0000} \textBF{84.0}} & {\color[HTML]{FE0000} \textBF{64.5}} & 6.2                                                                    & 1.8                                                                   & 272.7      \\ \bottomrule[1pt]                          
\end{tabular}
\vspace{-10pt}
\end{table*}

\textbf{Impact of varying numbers of ViT blocks in student models:}
To closely investigate the impact of varying numbers of ViT blocks on performance in student models, we train the AVTrack-MD-DeiT student model using a range of block counts from 4 to 8.
The evaluation results are shown in Table \ref{table_aba_vit_blocks}.
As observed, the number of ViT blocks in the student model directly affects both tracking performance and speed.
On average, increasing the number of ViT blocks leads to an upward trend in accuracy and model complexity but a decline in tracking speed.
As the number of blocks increases from 4 to 6, each additional block results in a greater than 1.0\% improvement in both average precision and average success rate, an increase of more than 0.4M in parameters, and a rise of over 0.1G in FLOPs, accompanied by an approximately 10\% decrease in speed.
When the number of blocks exceeds 6, further increases do not lead to significant performance gains, while model complexity continues to rise substantially and speed continues to decrease significantly.
Considering the balance between performance and speed, we have chosen to set the default number of ViT blocks in the student model to 6 in our implementation, which is half that of the teacher models.


\begin{table}[h]
\scriptsize
\setlength\tabcolsep{0.5pt} 
\centering
\caption{Comparison of different multi-teache knowledge distillation loss functions shows that the proposed MI maximization-based distillation framework outperforms the MSE loss.}
\label{table_MIkd_MSE}
\begin{tabular}{ccccccccccccccc}
\toprule[1pt]
\multirow{2}{*}{Method}          & \multirow{2}{*}{$\mathcal{L}_{MD}$} & \multicolumn{2}{c}{DTB70}     & \multicolumn{2}{c}{UAVDT}     & \multicolumn{2}{c}{VisDrone}  & \multicolumn{2}{c}{UAV123}    & \multicolumn{2}{c}{UAV123@10fps} & \multicolumn{2}{c}{Avg.}      \\
                                 &                         & Prec.         & Succ.         & Prec.         & Succ.         & Prec.         & Succ.         & Prec.         & Succ.         & Prec.           & Succ.          & Prec.         & Succ.       \\ \hline
Teacher                     & -                       & 84.3          & 65.0          & 82.1          & 58.7          & 86.0          & 65.3          & 84.8          & 66.8          & 83.2            & 65.8           & 84.1          & 64.4  \\ \hline
\multirow{2}{*}{Student} & MSE                     & 82.5          & 63.4          & 81.1          & 58.7          & 82.1          & 61.9          & 80.9          & 63.6          & 82.1            & 64.6           & 81.7          & 62.4   \\
                                 & JSD                     & \textBF{84.0} & \textBF{65.2} & \textBF{83.1} & \textBF{60.3} & \textBF{84.9} & \textBF{64.2} & \textBF{82.6} & \textBF{65.2} & \textBF{83.3}   & \textBF{65.5}  & \textBF{83.6} & \textBF{64.1} \\ \bottomrule[1pt]  
\end{tabular}
\end{table}

\textbf{Impact of Multi-Teacher Knowledge Distillation Based on MI Maximization:}
To demonstrate the advantage of our multi-teacher knowledge distillation through MI maximization, we train the model with both MSE loss and the proposed MI-based loss separately.
The evaluation results across five datasets are presented in Table \ref{table_MIkd_MSE}.
As observed, employing MSE loss as the distillation loss results in a more significant performance decline compared to our approach.
On average, employing the proposed $\mathcal{L}_{MD}$ results in a notable enhancement over the MSE loss, with improvements of 1.9\% in precision and 1.7\% in success rate, respectively.
Additionally, compared to the teacher model, our method incurs only a slight decrease of 0.5\% in average precision and 0.3\% in average success rate.
These comparisons show that the benefit of our multi-teacher knowledge distillation based on the MI maximization is that it can provide a more thorough measure of the relationship between features through MI-based loss, which enables the student model to learn a more effective representation from the teacher model.
On the other hand, our MI-based loss is less sensitive to noise and outliers compared to MSE, making it particularly effective in noisy environments.

\subsection{Qualitative Results:}

\begin{figure}[h]
\includegraphics[width=0.475\textwidth]{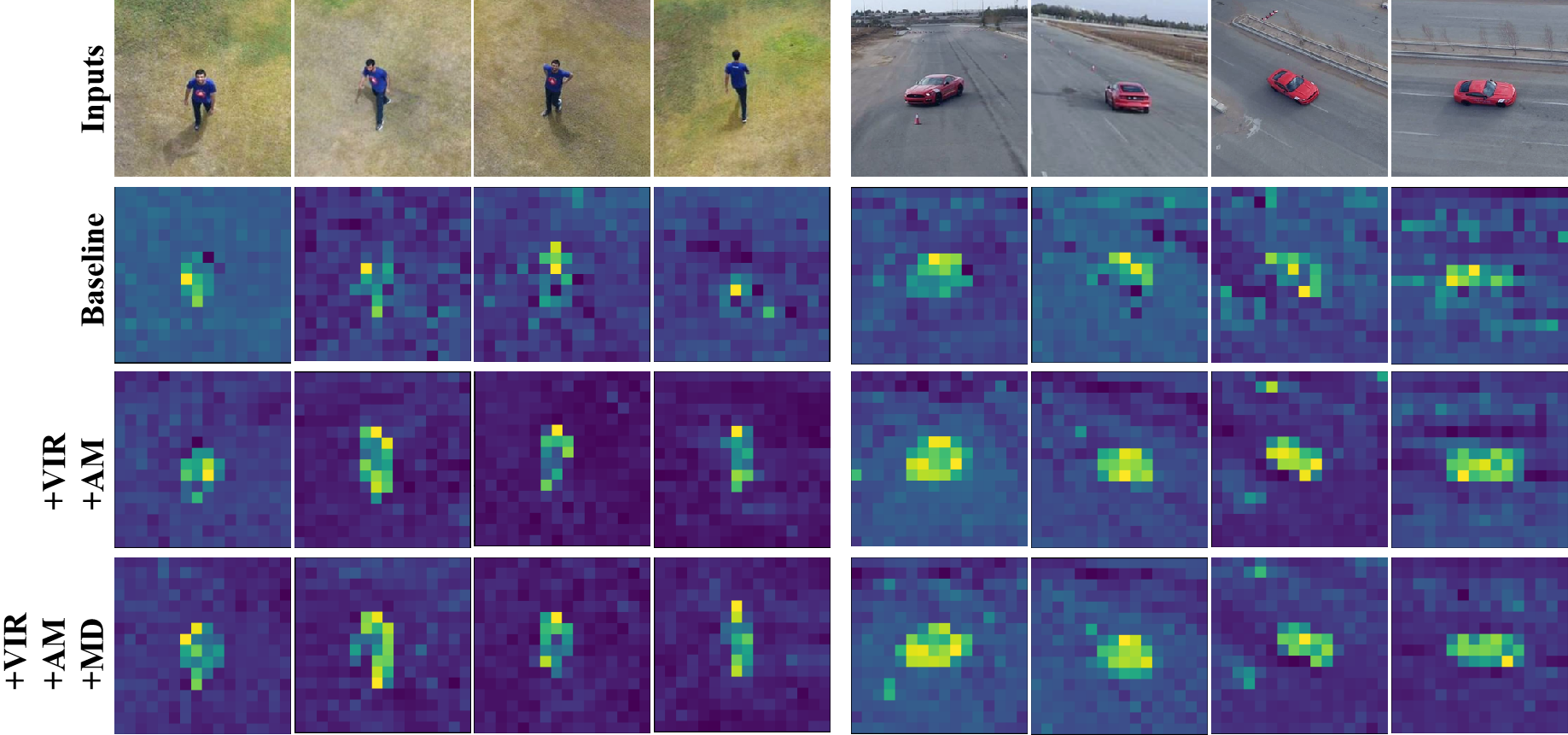}
\caption{For each group, we present input images from different viewpoints (top row), followed by the feature map generated by AVTrack-DeiT without the proposed VIR (second row), with the proposed VIR (third row), and with both the proposed VIR and MTKD (bottom row).}
\label{fig_featmap}
\end{figure}

\textbf{Visualization of feature maps:} 
In Fig. \ref{fig_featmap}, the first row presents the original target images from UAV123@10fps \cite{mueller2016benchmark}, while the corresponding feature maps generated by AVTrack-DeiT*, AVTrack-DeiT, and AVTrack-MD-DeiT are displayed in the second, third, and fourth rows, respectively.
Note that AVTrack-DeiT* refers to AVTrack-DeiT without the integration of the proposed VIR and AM components.
As observed, the feature maps generated by AVTrack-DeiT and AVTrack-MD-DeiT demonstrate greater consistency with changes in viewpoint, while the feature maps from AVTrack-DeiT* exhibit more significant variations, particularly at different viewing angles.
This suggests that the VIR component enhances the tracker's ability to focus on targets experiencing changes in viewpoint, thereby improving overall tracking performance. Furthermore, the proposed MD allows AVTrack-MD-DeiT to inherit this capability from AVTrack-DeiT, enabling it to maintain similar robustness to the teacher models while also enhancing model efficiency.
These qualitative results offer visual evidence of our method's effectiveness in learning view-invariant feature representations using ViTs.

\begin{figure}[h]
\centering
\includegraphics[width=0.475\textwidth]{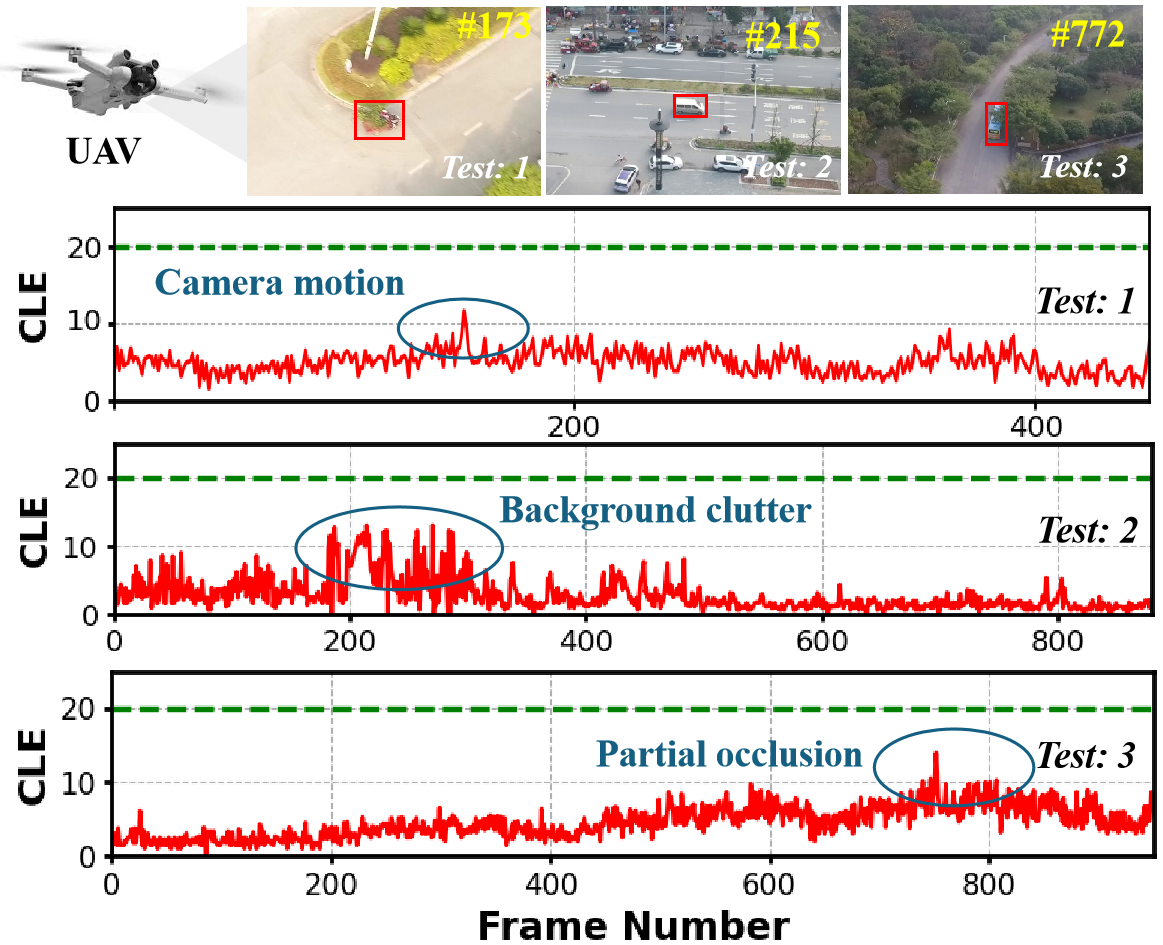}
\caption{Real-world UAV tracking test on an embedded device: acceptable tracking results defined by center location error (CLE) below 20.}
\label{real_world_test_fig}
\end{figure}

\subsection{Real-world Test}

To showcase the applicability of AVTrack-MD-DeiT for UAV tracking, we performed three real-world tests on the NVIDIA Jetson AGX Xavier 32GB.
As shown in Fig. \ref{real_world_test_fig}, the proposed tracker performs well in real-time despite challenges like camera motion, occlusion, and viewpoint changes, maintaining a Center Location Error (CLE) of under 20 pixels across all test frames, highlighting its high tracking precision.
Furthermore, AVTrack-MD-DeiT consistently achieves an average real-time speed of 46 FPS in tests, demonstrating its robustness and efficiency, making it ideal for UAV applications requiring high performance and speed.

\begin{figure}[t]
\centering
\includegraphics[width=0.475\textwidth]{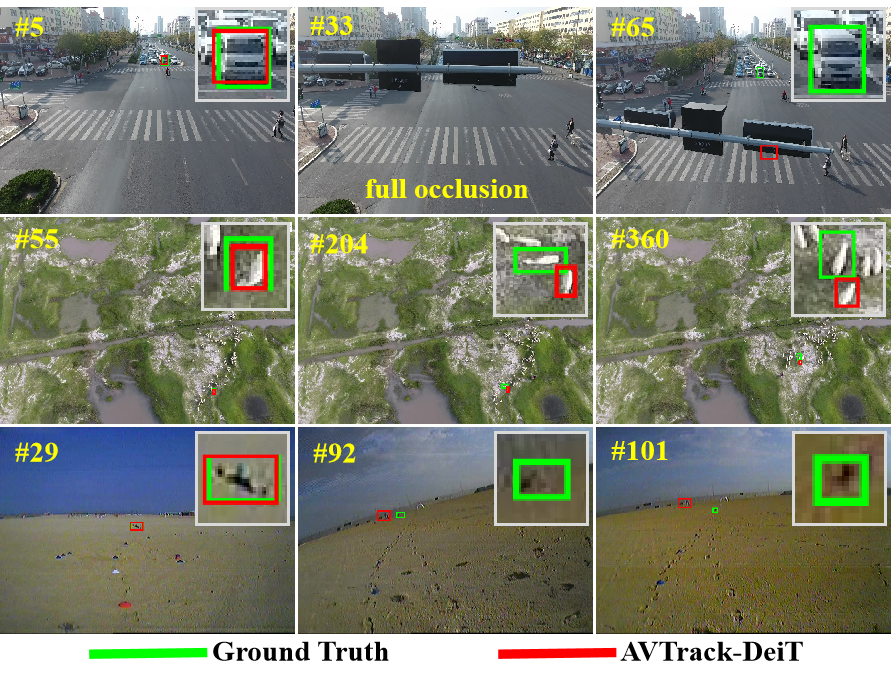}
\caption{Analysis of failure cases: AVTrack-DeiT struggles when the target is fully occluded, extremely small, or visually similar to surrounding objects, leading to tracking drift or loss.}
\label{fig_fail_case}
\end{figure}

\section{Limitations and future work}
\label{section_limitations}

\subsection{Limitations}

We present several representative failure cases in Fig. \ref{fig_fail_case}.
Although our tracker demonstrates strong robustness to viewpoint changes by learning view-invariant representations, it lacks a mechanism for reconstructing or retrieving target features after full occlusion (as shown in the first row), limiting its ability to maintain temporal continuity. Additionally, its performance degrades when tracking extremely small, visually similar, or camouflaged targets (as shown in the last two rows), mainly due to limited feature resolution and insufficient contrast between the target and the background, which impairs accurate localization and increases the risk of drift or tracking failure.

\subsection{Future work}

In future work, we plan to extend our framework by incorporating other advanced representation learning methods, such as contrastive learning and self-supervised feature alignment, to enhance robustness in the presence of full occlusion and background clutter. These techniques will help the tracker learn more discriminative and invariant features, improving its ability to differentiate the target from visually similar objects and maintain accurate tracking despite temporary loss of visual cues.

\section{Conclusions}\label{section_conclusion}
In this paper, we explore the effectiveness of a unified framework (AVTrack) for real-time UAV tracking through the use of efficient Vision Transformers (ViTs).
To achieve this, we introduced an adaptive computation paradigm that selectively activates transformer blocks. 
Additionally, to address the challenges of varying viewing angles commonly faced in UAV tracking, we utilized mutual information maximization to learn view-invariant representations.
Building on this, we have developed an improved AVTrack-MD model to enable more efficient UAV tracking by introducing a simple and effective MI maximization-based multi-teacher knowledge distillation (MD) framework.
Extensive experiments across six UAV tracking benchmarks show that AVTrack-MD not only delivers performance comparable to the AVTrack but also reduces model complexity, leading to a substantial improvement in tracking speed.

\bibliographystyle{IEEEtran}
\bibliography{main}

\end{document}